\pdfoutput=1
\documentclass[11pt]{article}

\usepackage{acl}

\usepackage{times}
\usepackage{latexsym}

\usepackage[T1]{fontenc}

\usepackage[utf8]{inputenc}

\usepackage{times}
\usepackage{latexsym}
\usepackage[T1]{fontenc}
\usepackage{soul}
\usepackage{multirow}
\usepackage{makecell}
\usepackage{boxedminipage}
\usepackage[utf8]{inputenc}
\usepackage{microtype}
\usepackage{graphicx}
\usepackage{soul}
\usepackage{float}
\usepackage{comment}
\usepackage{adjustbox}
\usepackage{booktabs}
\usepackage{amsmath}
\usepackage{mathabx}
\usepackage[shortlabels]{enumitem}
\usepackage[normalem]{ulem}
\usepackage{xspace}
\usepackage{array}
\usepackage{framed}
\usepackage{subcaption}
\usepackage{xcolor}
\usepackage{multicol}
\usepackage{algpseudocode}

\definecolor{purple}{rgb}{0.5,0,1}
\definecolor{dcyan}{rgb}{0.2,0.6,0.5}
\definecolor{darkgreen}{rgb}{0,200,0}
\definecolor{darkorange}{rgb}{138, 50, 0}
\definecolor{light-gray}{gray}{0.95} 
\definecolor{darkgreen}{RGB}{0,140,0}
\definecolor{darkred}{RGB}{200,0,0}
\definecolor{lightgreen}{RGB}{231,255,219}
\definecolor{lightred}{RGB}{252,231,234}
\definecolor{lightyellow}{RGB}{250,253,191}
\definecolor{DarkRed}{RGB}{130,25,0}
\newcommand{\hmt}{\textsc{Help me Think}\xspace}
\newif\ifsubmit
\submittrue
\ifsubmit
\newcommand{\elno}[1]{}
\else
\newcommand{\elno}[1]{{\color{magenta} elno: #1}}
\fi
\usepackage{algorithm}
\usepackage{algpseudocode}

\title{\hmt: A Simple Prompting Strategy for Non-experts to Create Customized Content with Models}

\author{Swaroop Mishra$^{2}$\thanks{~~Work done while interning at Microsoft Research.} $\;$ Elnaz Nouri$^{1}$ $\;$
\\\\
 $^1$Microsoft Research \; $^2$Arizona State University \;}

\begin{document}
\maketitle
\begin{abstract}

Controlling the text generated by language models and customizing the content has been a long-standing challenge. Existing prompting techniques proposed in pursuit of providing control are task-specific and lack generality; this provides overwhelming choices for non-expert users to find a suitable method for their task. The effort associated with those techniques, such as in writing examples, explanations, instructions, etc. further limits their adoption among non-expert users. In this paper, we propose a simple prompting strategy \hmt where we encourage large language models (such as GPT3 and ChatGPT) to help non-expert users by asking a set of relevant questions and leveraging user answers to execute the task. We demonstrate the efficacy of our technique \hmt on a variety of tasks. Specifically, we focus on tasks that are hard for average humans and require significant thinking to perform. We hope our work will encourage the development of unconventional ways to harness the power of large language models.


 
\end{abstract}

\section{Introduction}
\begin{figure*}[t]
    \centering
      \includegraphics[scale=0.58, trim=0cm 0cm 0cm 1.0cm,clip=false]{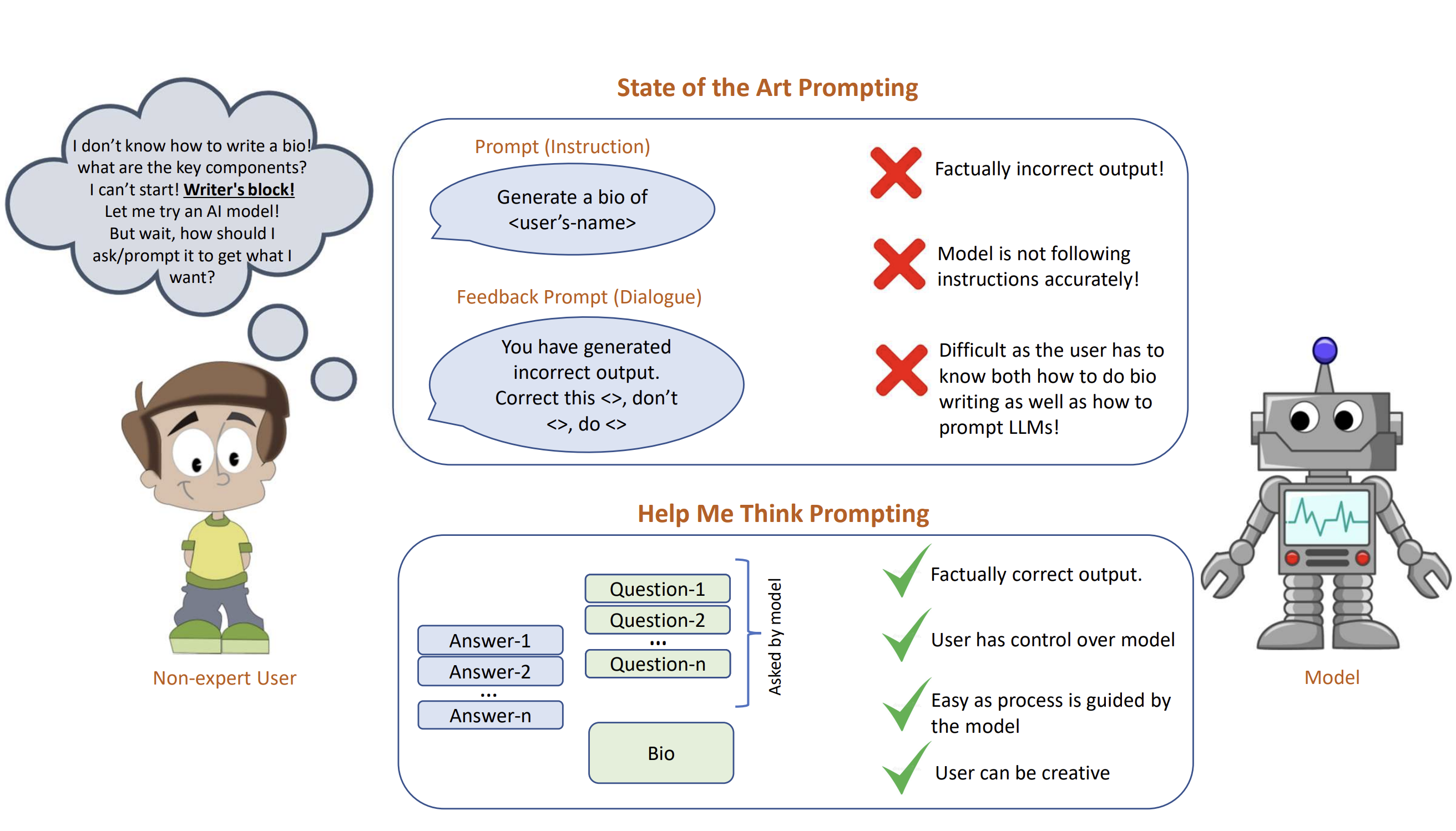}
    \caption{
 Illustration of \hmt prompting for a non-expert user on the bio generation task: detailed description of the example in Appendix ~\ref{sec:rel}
    }
    \label{fig:teaster}
\end{figure*}

Large language models (LLM) like GPT3, ChatGPT \cite{brown2020language} and PaLM \cite{chowdhery2022palm} have excelled in many NLP tasks, however creating customized content in the form of long text generation using these models is a challenge (Figure~\ref{fig:teaster}), as (1) models have not been reliable in following a list of instructions, (2) correcting model's output post generation by providing negative instructions (e.g. Don't do this) in the form of dialogue has also not worked consistently. More importantly, significant effort is required for non-expert users to write instructions containing the important ingredients of a task; for example, in order to create instructions for models to write the bio of a user, the user needs to think and provide various personal information. Travel \& event plan generation are similar to such tasks where a non-expert user has to think and find out various information necessary to make a plan, such as `number of attendees', `venue selection', `budget', `special arrangements', etc. 

We introduce \hmt: a simple prompting strategy for non-experts to write customized content with models. On a broader level, the goal of \hmt is similar to Make-A-Scene \cite{gafni2022make}; the application area, however, is vastly different as we focus on diverse applications that do not involve images and are purely based on text data. \hmt involves prompting models to ask relevant questions that reduce the thinking burden of users in identifying key information specific to that task. This application is in contrast to the dominant application of LLMs where various prompting techniques have been developed to enable LLMs 
in answering a question~\cite{liu2021pre}. We hypothesize that knowing the right question to ask is often more important than answering it, especially in the context of personalized content generation.


We experiment with six customized content generation tasks\footnote{We also explore an additional set of 57 tasks in Appendix~\ref{app:additional tasks}.}: (1) bio generation, (2) travel plan generation, (3) dialogue generation, (4) poem generation, (5) event summary generation, and (6) story generation. We prompt GPT3 and collect 68 questions corresponding to these tasks. We find that 100\% of the generated questions are valid and relevant. We use crowd-sourcing to collect answers; we leverage these question-answer pairs to prompt GPT3 again and get task-specific outputs e.g. bio, event plan, story, etc. With the \hmt prompting, we construct a dataset of $\sim$ 2k question-answer pairs with 180 task specific outputs spanning over the 6 tasks. We develop a questionnaire-based human evaluation scheme for crowdworkers to evaluate the quality of model generations: (1) questions and (2) task-specific outputs.

Our results show that 100\% of task-specific outputs generated by GPT3 are valid (e.g. valid bio) and $\sim$ 94\% of them do not contain any extra and irrelevant information. 
Moreover, we observe that, in $\sim$ 83\% of cases, GPT3 corrects typos/grammatical issues/invalid answers present in the crowdworkers' answers. GPT3 also adds appropriate context and generates coherent sentences for $\sim$ 99\% of cases. In 70\% of cases, GPT3 performs accurate knowledge transfer\footnote{With a tolerance level of 1 question-answer
pair for tasks with 4 questions (poem and dialogue tasks) and 2 question-answer pairs for tasks with more than 4 questions.} by transferring all information from input question-answer pairs to task-specific outputs. We hope the \hmt prompting and our focus on tasks hard for average humans will encourage the development of unconventional ways to harness the power of LLMs and bring more attention to the under-explored tasks.

\section{\hmt}
In this section, we introduce the \hmt prompting. We first present the algorithm behind \hmt. Next, we describe the prompts used in \hmt\footnote{Related work and detailed algorithm of \hmt are presented in Appendix~\ref{sec:relwork} and~\ref{sec:rel} respectively.} and explain the role of non-expert users in this prompting process.

\paragraph{Prompt:} Figure~\ref{fig:prompt1} illustrates the prompt given to the model that generates a question specific to the task of interest. Figure~\ref{fig:prompt2} shows that {\color{blue} model generates} a question and an answer, in response to the prompt. Figure ~\ref{fig:prompt3} shows how we generate multiple questions in this process. Figure ~\ref{fig:prompt4} shows where users fill in their answers. Figure~\ref{fig:prompt5} shows the prompt that is attached with the question-answer pairs which finally gives rise to the {\color{blue} model-generated task-specific output}.







\begin{figure}
\centering
\begin{boxedminipage}{\columnwidth}
\footnotesize
I am an expert \textit{\$task-executer\$}. I will ask some questions to collect information and then I will use the information to {\textit{\$do the task.\$}}\\
Question:
\end{boxedminipage}
\caption{Prompt given to the model to generate question. \textit{\$task-executer\$} and \textit{\$do the task.\$} are `bio generator' and `generate a bio for you' for the bio generation task. They vary across tasks.}
\label{fig:prompt1}
\end{figure}

\begin{figure}
\centering
\begin{boxedminipage}{\columnwidth}
\footnotesize 
I am an expert \textit{\$task-executer\$}. I will ask some questions to collect information and then I will use the information to {\textit{\$do the task.\$}}\\
Question: {\color{blue}{<model generates question>}\\
Answer: {<->}}
\end{boxedminipage}
\caption{{\color{blue} Model generation} in response to the prompt (Figure \ref{fig:prompt1}). Model also generates an answer along with the question, but {\color{blue} <->} indicates that we are not storing this information.}
\label{fig:prompt2}
\end{figure}
\begin{figure}
\centering
\begin{boxedminipage}{\columnwidth}
\footnotesize 
I am an expert \textit{\$task-executer\$}. I will ask some questions to collect information and then I will use the information to {\textit{\$do the task.\$}}\\
Question: {<model generated question>}\\
Answer: {<->}\\
Question: {\color{blue}{<model generated question>}\\
Answer: {<->}}\\
...
\end{boxedminipage}
\caption{Prompt (Figure \ref{fig:prompt1}) and the generated question-answer pair (Figure \ref{fig:prompt2}) are fed to the model to generate {\color{blue}new questions} for the task.}
\label{fig:prompt3}
\end{figure}

\begin{figure}[t]
\centering
\begin{boxedminipage}{\columnwidth}
\footnotesize 
I am an expert \textit{\$task-executer\$}. I will ask some questions to collect information and then I will use the information to {\textit{\$do the task.\$}}\\
Question: {<model generated question>}\\
Answer: {\color{red}{<user writes answer>}}\\
Question: {{<model generated question>}\\
Answer: {\color{red}{<user writes answer>}}}\\
...
\end{boxedminipage}
\caption{User writes {\color{red} answers} to the questions (Figure ~\ref{fig:prompt4}) generated by model}
\label{fig:prompt4}
\end{figure}

\begin{figure}
\centering
\begin{boxedminipage}{\columnwidth}
\footnotesize 
I am an expert \textit{\$task-executer\$}. I will ask some questions to collect information and then I will use the information to {\textit{\$do the task.\$}}\\
Question: {<model generated question>}\\
Answer: {{user written answer}>}\\
Question: {{<model generated question>}\\
Answer: {{<user written answer}>}}\\
...\\
Write a \textit{\$task-specific-output\$} using the questions and answers above. \textit{\$task-specific-instruction\$}\\
{\color{blue}{<model generates task-specific-output>}}
\end{boxedminipage}
\caption{A task-specific prompt is added after the model generated question-answer pairs. {\color{blue}{<model generates task-specific-output>}} in response to the prompt.  \textit{\$task-specific-output\$} for the bio generation task is `a long bio about John'. \textit{\$task-specific-instruction\$} is optional, e.g. `Introduce names to represent characters.' for the story generation task.}
\label{fig:prompt5}
\end{figure}

\section{Experiments}
We have conducted experiments with \hmt on a diverse set of novel tasks. In this section, we describe these tasks and provide the details of the data collection and our evaluation setup.

\subsection{Tasks}\label{subsec:tasks}
We experiment with six customized content generation tasks \footnote{Examples of each of the tasks are in Appendix~\ref{sec: userinput and gpt3output}}: (1) bio generation, (2) travel plan generation, (3) dialogue generation, (4) poem generation, (5) event summary generation and (6) story generation.
\subsection{Data Collection}
We prompt GPT3 (Figure~\ref{fig:prompt3}) to generate task-specific questions automatically. We set up a crowdsourcing task to get answers to the questions. We use a private group of crowdworkers that leverages a set of internal tools specific to the anonymous organization; they are instructed to write diverse outputs while collecting answers \footnote{More details of crowdsourcing is in Appendix~\ref{sec:crowdsourcing}}. 
GPT3 is prompted with question-answer pairs and a task-specific prompt (Figure ~\ref{fig:prompt5}) to generate task-specific outputs. 
\subsection{Statistics}
We collect a total of $\sim 2k$ QA pairs. Table~\ref{tab:stastics} shows some key statistics in our collected data.

\begin{table}
    \begin{tabular}{lc}
        \toprule
        category  & \# of instances  \\
        \midrule
        {task} & 6 \\ 
        {questions} & 68\\
        {question-answer pair} &  2040 \\ 
        {task outputs}   & 180 \\ 
        \bottomrule
    \end{tabular}
    \caption{Key statistics of our collected data. 
    }
    \label{tab:stastics}
\end{table}
\subsection{Evaluation}
We use human evaluation since content generation tasks are open-ended and are hard to be captured by automated evaluation metrics. Each question associated with each task is evaluated by 3 annotators. The annotators evaluate various aspects of generated text by answering the associated questions.
\begin{table*}
\centering
    \begin{tabular}{lcccccc|c}
        \toprule
        category  & bio & travel plan & dialogue & poem & event summary & story & avg.  \\
        \midrule
        {Validity}   & 100 & 100 &  100& 100 & 100& 100 &  100\\ 
        {Relevance} & 100 & 100  & 100 & 100 & 100  & 100 & 100\\
        \bottomrule
    \end{tabular}
    \caption{Evaluation (majority voting of 3 annotators) of model generated questions for each task.
    }
    \label{tab:questioneval}
\end{table*}

\begin{table*}
    \begin{tabular}{lcccccc|c}
        \toprule
        category  & bio & travel plan & dialogue & poem & event summary & story & avg.  \\
        \midrule
        {Validity} & 100 & 100 & 100 & 100 & 100 & 100 & 100 \\ 
        {Knowledge Absorption} & 86.66 & 3.33 & 70 & 90 & 86.66 & 83.33 & 70\\
        {Relevance} &  100 & 93.33 & 76.67 & 96.67 & 96.67 & 100 & 93.89\\ 
        {Robustness}   & 96.67 & 50 & 71.42 & 100 & 77.78 & 100 & 82.65 \\ 
        {Coherence} & 100 & 100 & 96.15 & 100 & 100 & 100 & 99.36\\  
        \bottomrule
    \end{tabular}
    \caption{Model performance on different evaluation aspects for each task (majority voting of 3 annotators). For the `Knowledge Absorption' category, a tolerance level of 1 question-answer pair for tasks with 4 questions (poem and dialogue tasks) and 2 question-answer pairs for tasks with more than 4 questions are taken.
    }
    \label{tab:main result}
\end{table*}
\subsubsection{Evaluation of Model Generated Questions}
The annotators evaluate the following two aspects of the questions generated by models. Each question is evaluated by three annotators.\\

\noindent\textbf{Validity} \textit{Is it a question? (and not a statement or any other text.)}\\
\textbf{Relevance} \textit{Is it relevant to the underlying task?} E.g. bio, travel plan, story, etc.  \\

\subsubsection{Evaluation of Model Generated Task-specific Outputs:}
Furthermore, each task-specific output (e.g. bio, story, etc.) generated by GPT3 and its corresponding input (generated by GPT3 and answered by the user) is evaluated by three annotators. Each annotator is asked to answer a question that covers a specific part of the evaluation as follows:\\

\noindent\textbf{Validity}: \textit{Is the output a valid task-specific output?} E.g. is it a valid bio? (for the bio-generation task).\\
\textbf{Knowledge Absorption}: \textit{Does the output incorporate all the facts and information from the input?}\\
\textbf{Relevancy}: \textit{Does the output have unrelated information that is not present in the input?}\\
\textbf{Robustness}: \textit{Has the output fixed any typos or grammatical errors or invalid answers present in the input, instead of copying the same?}\\
\textbf{Coherence}: \textit{Can you find any example of output having additional related and contextual information written as a coherent sentence in addition to what was already present in the input?}

\section{Results}
We report the task-wise performance of GPT3 and analyze its variation across different aspects of evaluation.

\paragraph{Insights:}
\textit{We observe that 100\% of the questions generated by GPT3 are valid and relevant to the task (Table ~\ref{tab:questioneval}).} Table ~\ref{tab:main result} shows the performance of GPT3 in producing task-specific outputs. We observe that \textit{(1) 100\% of the generations are valid task-specific outputs, (2) 93.89\% of the generations do not contain irrelevant content, and (3) 99.36\% times, GPT3 improves the coherence of text over the information presented to it in the form of input-output examples. (4) 82.65\% times, GPT3 fixes typo/grammatical issues present in the user-written answers.} However, we see that the knowledge absorption is low (70\%), which we analyze further.


We also ask crowdworkers to write explanations; we analyze those to better understand the knowledge absorption in the task specific output generated by GPT3. We understand that for certain tasks like travel plan generations, some question-answer pairs are not important; they are not always necessary to be part of the plan\footnote{Additional evaluation and analysis are in Appendix~\ref{sec:adanalysis}.}. 

\paragraph{Extension to Other Tasks:}
We apply \hmt on 57 additional tasks (Appendix ~\ref{app:additional tasks}) and find that \hmt is effective in generating valid and relevant questions. 
This shows the generalization of \hmt beyond the six tasks we have analyzed (Section ~\ref{subsec:tasks}).

\paragraph{Qualitative Comparison to Conversational Language Models:}
  \hmt can also be used for customized content creation with conversational models such as ChatGPT and Bard. Here the "role" parameter can be used as part of the prompts, specifically the prompts in Figure ~\ref{fig:prompt1}-~\ref{fig:prompt5} can now be reframed from first person to second person since these models are designed to interact and respond to instructions. For example, the prompt in Figure ~\ref{fig:prompt1} can now be "You are an expert \textit{\$task-executer\$}. You will ask some questions to collect information and then You will use the information to {\textit{\$do the task.\$}}."
  In our qualitative analysis without \hmt\footnote{See Appendix~\ref{sec:conversational} for examples.}, conversational language models does not show consistent question answering behavior and hallucination is still a major problem. \hmt fixes this issue along with the knowledge cut off issue of some conversational language models because it collects facts from the user during the interaction. The instruction following capability of conversational language models is better than conventional language models, but the number of instructions needed to correct/customize its generations will probably be higher in regular multi-turn conversations compared to Help-Me-Think. We view Help-Me-Think as a model agnostic approach that can be used and adopted with any of the state of the art models. 
 
\section{Conclusion}
We introduce \hmt to help non-expert users prompt models for the relatively under-explored customized content generation tasks. We demonstrate the efficacy of \hmt on 6 different tasks. Our results show that (1) questions generated by GPT3 are valid and relevant, and (2) task-specific outputs generated by GPT3 are valid, relevant, robust, coherent, and have significant knowledge absorption. We hope this will bring more attention to the development of unconventional applications of LLMs in helping humans perform the tasks that are hard for non-expert users, in contrast to the conventional tasks like question-answering where models chase human baseline. 

\section{Limitation}
\hmt is a model agnostic approach that allows users to inject facts to accomplish tasks with a variety of large language models through simple Q\&A but additional experiments are needed to establish its effectiveness on new language models with different training paradigms and capabilities. 
The entire study is conducted only with tasks of English language. Expanding the scope of \hmt to other languages will increase the scope for non-expert users. A large scale evaluation setup is further needed to reach \hmt to non-expert users.





\bibliography{anthology,custom}
\bibliographystyle{acl_natbib}
\clearpage
\appendix


\section{Related Work}\label{sec:relwork}
\paragraph{Prompting and Learning from Instructions:}
The success of large language models~\cite{brown2020language, chowdhery2022palm} has empowered the development of various prompting techniques~\cite{liu2021pre}. Instructions, proposed as an extension to prompts, describe tasks in natural language~\cite{efrat2020turking, weller2020learning} and guide models to generalize to unseen tasks~\cite{mishra2022cross,wei2021finetuned, ouyang2022training, sanh2021multitask, zhong2021adapting, wang2022self2, scaria2023instructabsa} without requiring task-specific training. Prompts and Instructions are shown to be helpful in low-resource settings~\cite{le2021many,puri2022many}. Several variants of prompting such as chain of thought~\cite{wei2022chain} or scratchpad~\cite{nye2021show}, majority voting~\cite{wang2022self}, reframing~\cite{mishra-etal-2022-reframing}, least-to-most prompting~\cite{zhou2022least}, question decomposition~\cite{, khot2020text, patel2022question} have been shown to be effective across various tasks. Efficacy of the Prompting/Learning from Instruction techniques has been shown across diverse applications~\cite{wang2022benchmarking} such as dialog~\cite{gupta2022improving}, NER~\cite{wang2022instructionner}, program synthesis~\cite{kuznia2022less}, style transfer~\cite{reif2021recipe}, tabular question answering~\cite{luo2022biotabqa}, relation extraction~\cite{chen2021adaprompt}, biomedical applications~\cite{parmar2022boxbart}. In contrast to prior works, we (1) focus on a diverse set of creative tasks as our application area, (2) build a non-expert user-centric technique (3) leverage language models for assisting users to think by asking questions (instead of just answering questions posed by humans), and in this process engage users that subsequently helps them learn the thinking process.



\paragraph{Creative Tasks using GPT3:}
GPT3 has been recently used for creative tasks such as writing a paper\footnote{https://www.scientificamerican.com/article/we-asked-gpt-3-to-write-an-academic-paper-about-itself-then-we-tried-to-get-it-published/, https://hal.archives-ouvertes.fr/hal-03701250/document}, poem\footnote{https://www.newyorker.com/culture/culture-desk/the-new-poem-making-machinery
}, article\footnote{https://www.theguardian.com/commentisfree/2020/sep/08/\\robot-wrote-this-article-gpt-3}, and book\footnote{https://medium.com/swlh/i-wrote-a-book-with-gpt-3-ai-in-24-hours-and-got-it-published-93cf3c96f120, https://augmentedrobot.medium.com/281-years-in-the-making-gpt-3-and-de-la-mettrie-de03eea374e4}. However, controlling model generation in these creative tasks is a challenge. On a broader level, controlling content in text generated by pretrained language models has been a challenge in NLP \cite{perez2022finding}. The \hmt framework provides a model-guided approach by generating and asking questions to assist users in thinking through various steps and at the same time controlling the content of the text generated by the model.
We further prove the efficacy of the \hmt approach by applying it to a set of 63 diverse tasks. These tasks are novel and creative and can potentially form a benchmark dataset for future research.

\paragraph{Interactive Learning}
Interactive question answering has been utilized in several recent works~\cite{activeruqi, zhao2022compositional, yao-etal-2020-imitation} around semantic parsing. In these cases, the target output is code and the task is being decomposed into smaller sub-tasks. A similar approach has also been applied to the robotics domain e.g. SayCan~\cite{ahn2022can}. \hmt is different in two ways (1) the schema in case of semantic parsing tasks and the set of actions in robotics tasks are fixed, whereas in \hmt, schema is dynamic and is derived from a high level description of the task (2) in the approaches for semantic parsing and robotics tasks, there is a concern if the generated output can be executed on downstream tasks (e.g. in guiding a robot), however that concern is relaxed in \hmt as the output itself is in natural language.

\section{\hmt algorithm and Description}\label{sec:rel}
\paragraph{Algorithm:} Algorithm~\ref{alg:hmt} illustrates the detailed algorithm behind \hmt. It has 3 stages: Stage 1 (Generate Questions), Stage 2 (Collect Answers) and Stage 3 (Generate Task-specific Output). We describe each stage in detail below.

\paragraph{Description}
In figure~\ref{alg:hmt}, a non-expert user is asked to write a biography, but this is a hard task for the user since it demands thinking about key and necessary components for a biography which the user might not know about or the user might just simply be dealing with writer's block when faced with creative writing tasks. The user decides to get help from an AI model. The user wants to try to prompt an AI model (using state-of-the-art instruction prompting paradigms), but the model produces factually incorrect output for him. Next, the user tries to interact with the model and provide feedback to correct the model prediction (dialogue paradigm), but this approach is also a failure because it is a challenge for models to accurately follow feedback instructions. For the majority of non-expert users, figuring out an effective prompting strategy is a major challenge. Finally, \hmt helps the user generate the factually correct biography via the model by guiding the user in the process by asking questions, this alleviates the cognitive demand on the user significantly. By removing the hurdles out of the way of the user in writing his biography, \hmt also allows the user to take a step further and focus on creativity and quality.
\begin{algorithm}
\caption{\hmt algorithm}\label{alg:hmt}
\begin{algorithmic}[1]
\State Generate question-generation prompt by replacing task-specific variables in the prompt (figure ~\ref{fig:prompt1}) \Comment Start of Stage 1 (Generate Questions)
\State Setup stop condition (e.g. ?, ‘Answer:’, newline ) and ask GPT3 to complete with ‘Question: ' prompt 
\State Repeat Step 2 until the model starts generating repetitive, redundant, or irrelevant content.
\State If default settings are not producing the expected output, then try these (1) control temperature, (2) control max output size, (3) add additional task-specific instruction or (4) add an example question. \Comment End of Stage 1 (Generate Questions)
\State Ask user to answer each of the questions and pose customization requirements (Figure~\ref{fig:prompt4}) \Comment Stage 2 (Collect Answers)
\State Generate task-specific output generation prompt by replacing task-specific variables in the prompt (figure~\ref{fig:prompt5}) \Comment Start of Stage 3 (Generate Task-specific Output)
\State If question-answer pairs are dependent on each other, collect all of them to feed them all at once to model, else feed them batch by batch 
\State Setup appropriate stop condition and ask GPT3 to complete output after the prompt 
\State Concatenate task-specific output if step 6 was done in batches 
\State If default settings are not producing valid output, then try these (1) control temperature, (2) control max output size or (3) add additional task-specific instruction.  \Comment End of Stage 3 (Generate Task-specific Output)

\end{algorithmic}
\end{algorithm}
\paragraph{Role of Non-expert User:} Figure ~\ref{fig:teaster} illustrates the role of a non-context user in performing bio generation tasks. It requires a lot of thinking for the non-expert user to write a bio as it demands the identification of key ingredients corresponding to the bio generation task. \hmt in contrast to other prompting techniques helps the user perform the task via the model with minimal effort, as they just need to answer the questions generated by the model (Figure ~\ref{fig:prompt4}).

\section{Detailed Prompts and Hyperparameters}
In this section, we describe the initial prompts (Figure~\ref{fig:prompt1}) we use across tasks. Figures~\ref{fig:prompt1bio}, \ref{fig:prompt1travel}, \ref{fig:prompt1dialogue}, \ref{fig:prompt1poem}, \ref{fig:prompt1event}, \ref{fig:prompt1story} show the prompts for bio, travel plan, dialogue, poem, event summary and story generation tasks respectively.

Figure~\ref{tab:allquestions} illustrates all questions generated by GPT3 for various tasks in response to our prompting (Figure~\ref{fig:prompt2},\ref{fig:prompt3}).

Figure~\ref{fig:prompt2bio}, \ref{fig:prompt2travel}, \ref{fig:prompt2dialogue}, \ref{fig:prompt2poem}, \ref{fig:prompt2event}, \ref{fig:prompt2story} show the prompt used to generate task-specific output from GPT3 (as in figure~\ref{fig:prompt5}).

We use the following hyper-parameters while querying GPT3 for various tasks: \texttt{engine=text-davinci-002}, \texttt{temperature=0.7}, \texttt{max\_tokens=512},\texttt{top\_p=1},
\texttt{frequency\_penalty=0}, \texttt{presence\_penalty=0}.

\begin{figure}[H]
\centering
\begin{boxedminipage}{\columnwidth}
\footnotesize
I am an expert in generating Bio of people. I ask questions to gather information. Then I use these information to generate bio.\\
Question:
\end{boxedminipage}
\caption{Prompt given to model to generate question about the bio generation task.}
\label{fig:prompt1bio}
\end{figure}
\begin{figure}[H]
\centering
\begin{boxedminipage}{\columnwidth}
\footnotesize
I am a famous travel planner. I will ask clarifying question to collect information and then I will write an awesome travel plan and schedule for you.\\
Question:
\end{boxedminipage}
\caption{Prompt given to model to generate question about the travel plan generation task.}
\label{fig:prompt1travel}
\end{figure}
\begin{figure}[H]
\centering
\begin{boxedminipage}{\columnwidth}
\footnotesize
I am a famous dialogue writer. I will ask simple questions to collect information and then I will write a dialogue series specially for you.\\
Question:
\end{boxedminipage}
\caption{Prompt given to model to generate question about the dialogue generation task.}
\label{fig:prompt1dialogue}
\end{figure}
\begin{figure}[H]
\centering
\begin{boxedminipage}{\columnwidth}
\footnotesize
I am a famous poet. I will ask clarifying question to collect information and then I will write a poem.\\
Question:
\end{boxedminipage}
\caption{Prompt given to model to generate question about the poem generation task.}
\label{fig:prompt1poem}
\end{figure}
\begin{figure}
\centering
\begin{boxedminipage}{\columnwidth}
\footnotesize
I am a famous event planner. I will ask clarifying question to collect information and then I will write an awesome event plan for you.\\
Question:
\end{boxedminipage}
\caption{Prompt given to model to generate question about the event summary generation task.}
\label{fig:prompt1event}
\end{figure}
\begin{figure}
\centering
\begin{boxedminipage}{\columnwidth}
\footnotesize
I am an expert script writer. I will ask some simple questions to collect information and then I will write a story of your choice.\\
Question:
\end{boxedminipage}
\caption{Prompt given to model to generate question about the story generation task.}
\label{fig:prompt1story}
\end{figure}

\begin{table*}
    \centering
    \resizebox{\textwidth}{!}{
    \begin{tabular}{cl|cl}
        \toprule
        task & questions& task & questions  \\
        \midrule
         \multirow{32}{*}{\makecell{bio\\ }} & What do you do? &  \multirow{8}{*}{\makecell{story}} & What type of story would you like me to write?\\ 
      	& What are your interests? & & What is the main plot of your story?\\	
      	& What are your skills? & & What is the ending of your story?\\	
      	& What are your experiences? & & What are the main characters in your story?\\
        & What is your education? & &Where does the story take place?\\	
        & What is your work history?  & &Why does the story end the way it does?\\	
        & What are your awards? & &What is the main conflict in your story?\\	
        & What is your family background? & & What message do you want your story to send?\\
        & What are your hobbies? & & -------------------------------------------------------\\
        & What is your favorite thing to do? & \multirow{12}{*}{\makecell{event\\ plan}} &  What type of event are you looking to plan?\\	
        & What is your favorite food? & & What is the purpose of the event?\\	
        & What is your favorite color?& & Who is the target audience for the event?\\
        & What is your favorite animal? & & When is the event taking place?\\
        & What is your favorite sport? & &  Where is the event taking place?\\	
        & What is your favorite team? & & What is the budget for the event?\\	
        & What is your favorite movie? & & What is the expected headcount for the event?\\	
        & What is your favorite book? & & What is the theme of the event?\\
        & What is your favorite music? & & What activities do you want to include in the event?\\	
        & What is your favorite TV show? &  & Do you have any specific requests for the event?\\	
        & What is your favorite vacation spot? & & What is the timeline for the event?\\	
        & What is your favorite thing to do on a weekend?& & What is the expected outcome of the event?\\	
        & What is your favorite thing to wear? & &------------------------------------------------------\\	
        & What is your favorite thing to do for fun? & \multirow{4}{*}{\makecell{dialogue}} & What is the most important thing in your life?\\	
        & What is your favorite thing to do with friends? & & What are your hopes and dreams for the future?\\	
        & What is your favorite thing to do alone? & & What makes you happy?\\	
        & What is your favorite place to go? & & What is your favorite thing about life?\\	
        & What is your favorite thing to do on a date?& &------------------------------------------------------\\	
        & What is your favorite thing to do when you’re feeling down? & \multirow{8}{*}{\makecell{travel\\ plan}} & How many people are in your party?\\	
        & What is your favorite thing to do when you’re happy? & & What are the ages of the members of your party?\\
        & What is your favorite thing to do when you’re bored? & & What is the budget for your trip?\\	
        & What is your favorite thing to do when you’re stressed? & & What are your preferred travel dates?\\
        
        & What is your favorite thing to do when you’re tired? & & What is your preferred mode of transportation? \\
        &------------------------------------------------------& & What are your preferred accommodation options?\\	
         \multirow{4}{*}{\makecell{poem}} & What is the occasion? & & What are your preferred activities while on vacation?\\	
         & What is the mood?& & What are your preferred food options while on vacation?\\
         & What is the theme? & \\	
         & What is the tone? &\\
        \bottomrule \\
    \end{tabular}}
    \caption{
        Questions generated by GPT3 for various tasks in response to our prompting (Figure~\ref{fig:prompt2},\ref{fig:prompt3}).
    }
    \label{tab:allquestions}
\end{table*}

\begin{figure}
\centering
\begin{boxedminipage}{\columnwidth}
\footnotesize
I am an expert in generating Bio of people. I ask questions to gather information. Then I use these information to generate bio.\\
Question: {<model generated question>}\\
Answer: {{user written answer}>}\\
Question: {{<model generated question>}\\
Answer: {{<user written answer}>}}\\
...\\
Write a long bio about John using the questions and his answers above.\\
{\color{blue}{<model generates task-specific-output>}}
\end{boxedminipage}
\caption{Prompt given to model to generate task-specific output about the bio generation task.}
\label{fig:prompt2bio}
\end{figure}
\begin{figure}
\centering
\begin{boxedminipage}{\columnwidth}
\footnotesize
I am a famous travel planner. I will ask clarifying question to collect information and then I will write an awesome travel plan and schedule for you.\\
Question: {<model generated question>}\\
Answer: {{user written answer}>}\\
Question: {{<model generated question>}\\
Answer: {{<user written answer}>}}\\
...\\
Based on the information provided, I would recommend the following travel schedule and budget for your trip:\\
{\color{blue}{<model generates task-specific-output>}}
\end{boxedminipage}
\caption{Prompt given to model to generate task-specific output about the travel plan generation task.}
\label{fig:prompt2travel}
\end{figure}
\begin{figure}
\centering
\begin{boxedminipage}{\columnwidth}
\footnotesize
I am a famous dialogue writer. I will ask simple questions to collect information and then I will write a dialogue series specially for you.\\
Question: {<model generated question>}\\
Answer: {{user written answer}>}\\
Question: {{<model generated question>}\\
Answer: {{<user written answer}>}}\\
...\\
Write a nice dialogue series using all the questions and answers above.\\
Person 1:\\
{\color{blue}{<model generates task-specific-output>}}
\end{boxedminipage}
\caption{Prompt given to model to generate task-specific output about the dialogue generation task.}
\label{fig:prompt2dialogue}
\end{figure}
\begin{figure}
\centering
\begin{boxedminipage}{\columnwidth}
\footnotesize
I am a famous poet. I will ask clarifying question to collect information and then I will write a poem.\\
Question: {<model generated question>}\\
Answer: {{user written answer}>}\\
Question: {{<model generated question>}\\
Answer: {{<user written answer}>}}\\
...\\
Write a poem using the question and answers above.\\
{\color{blue}{<model generates task-specific-output>}}
\end{boxedminipage}
\caption{Prompt given to model to generate task-specific output about the poem generation task.}
\label{fig:prompt2poem}
\end{figure}
\begin{figure}
\centering
\begin{boxedminipage}{\columnwidth}
\footnotesize
I am a famous event planner. I will ask clarifying question to collect information and then I will write an awesome event plan for you.\\
Question: {<model generated question>}\\
Answer: {{user written answer}>}\\
Question: {{<model generated question>}\\
Answer: {{<user written answer}>}}\\
...\\
Write a detailed event plan based on the questions and answers above.\\
{\color{blue}{<model generates task-specific-output>}}
\end{boxedminipage}
\caption{Prompt given to model to generate task-specific output about the event summary generation task.}
\label{fig:prompt2event}
\end{figure}
\begin{figure}
\centering
\begin{boxedminipage}{\columnwidth}
\footnotesize
I am an expert script writer. I will ask some simple questions to collect information and then I will write a story of your choice.\\
Question: {<model generated question>}\\
Answer: {{user written answer}>}\\
Question: {{<model generated question>}\\
Answer: {{<user written answer}>}}\\
...\\
Write a long story using the questions and answers above. Introduce names to represent characters.\\
{\color{blue}{<model generates task-specific-output>}}
\end{boxedminipage}
\caption{Prompt given to model to generate task-specific output about the story generation task.}
\label{fig:prompt2story}
\end{figure}

\section{User Inputs and GPT3 Outputs}\label{sec: userinput and gpt3output}
For each of the tasks, we illustrate a sample user input and GPT3 output.

Table~\ref{tab:biotaskto}, \ref{tab:traveltaskto}, \ref{tab:dialoguetaskto}, \ref{tab:poemtaskto}, \ref{tab:eventsummarytaskto}, \ref{tab:storytaskto} shows sampler user input and task-specific output generated by GPT3 for bio, travel plan, dialogue, poem, event summary and story generation task respectively.

\begin{table*}
    \footnotesize 
    \resizebox{\textwidth}{!}{
    \centering
    \begin{tabular}{p{8mm}p{4 cm}p{2.3cm}|p{7cm}}
        \toprule
        task & questions & answers & task-specific outputs \\
        \midrule
        \multirow{54}{*}{\makecell{bio\\ }}&What are your hobbies? & Cooking & \multirow{4}{=}{\setlength\parskip{\baselineskip}%
        John is a avid hobbyist who loves to cook and ride bikes. His favorite food is pizza, and his favorite color is black. John is a very friendly person who loves to meet new people. He is a very outgoing person and loves to have fun. John is a very hard worker and is always looking to improve himself. He is a very dedicated person and is always willing to help others.\newline\newline John is a highly accomplished individual who has made a significant impact in his field. He holds a Bachelor of Commerce from a prestigious university and has worked his way up through the ranks to become a Junior Executive. John has been recognised for his outstanding work with numerous awards, including Employee of the Month. John comes from a strong family background; his family belongs to South Indian origin and he is extremely proud of his heritage. John is a highly motivated individual who always strives to achieve the best possible results. He is a true asset to any organisation and has a bright future ahead of him.\newline\newline John is a fun-loving guy who loves spending time with his friends. He enjoys doing things like going out for a day, mimicry, and just having a good time. He loves to dress up and look his best, and he enjoys wearing dresses and other fun clothes. When it comes to fun, John is definitely the life of the party!\newline\newline John is a fun-loving guy who loves to dance. When he's feeling down, he likes to rest up and when he's feeling up, he loves to go tubing. Rome is his favorite place to go and he loves to explore new places.\newline\newline John is a voracious reader, and his favorite book is To Kill a Mockingbird. He loves all genres of music, but his favorite is As it Was. He is a huge fan of The Boys, and his favorite vacation spot is Bali, Indonesia.\newline\newline John is a huge animal lover, and his favorite animal is a dog. He's also a big fan of cricket, and his favorite team is Real Madrid. He loves watching Bhool Bhulaiyaa, and it's one of his favorite movies.\newline\newline John is an avid exerciser who loves to stay active when he's feeling happy. When bored, he enjoys working on puzzles to keep his mind sharp. And when feeling stressed, he finds that some breathing exercises help him to relax. Yoga is one of his favorite things to do when tired, as it helps him to stretch and wind down after a long day.\newline\newline John is a writer who is interested in trekking and has great time management skills. He has been promoted in the past and is looking to continue his writing career.}\\
        &What is your favorite thing to do? & Riding bikes & \\
        &What is your favorite food? & Pizza & \\
        &What is your favorite color?& Black& \\
        &What is your education?& Bachelor of Commerce & \\
        &What is your work history? & Junior Executive\\
        &What are your awards?& Employee of the month\\
        &What is your family background?& Family belongs to South Indian origin\\
        &What is your favorite thing to do on a weekend? & Day out with friends\\
        &What is your favorite thing to wear?& Dress\\
        &What is your favorite thing to do for fun?&Mimicry\\
        &What is your favorite thing to do with friends?& Outing\\
        &What is your favorite thing to do alone?& Dance\\
        &What is your favorite place to go?& Rome, Italy\\
        &What is your favorite thing to do on a date?& Go tubing\\
        &What is your favorite thing to do when you're feeling down?&Rest up\\
        &What is your favorite book?& To Kill a Mockingbird\\
        &What is your favorite music?& As it was\\
        &What is your favorite TV show?&The Boys\\
        &What is your favorite vacation spot?& Bali, Indonesia\\
        &What is your favorite animal?& Dog\\
        &What is your favorite sport?& Cricket\\
        &What is your favorite team?& Real Madrid, Football\\
        &What is your favorite movie? & Bhool Bhulaiyaa\\
        &What is your favorite thing to do when you're happy?&Exercise\\
        &What is your favorite thing to do when you're bored?& Work on Puzzle\\
        &What is your favorite thing to do when you're stressed?& Do some breathing exercise\\
        &What is your favorite thing to do when you're tired?& Yoga\\
        &What do you do?&Write\\
        &What are your interests?&Trekking\\
        &What are your skills?&Time management\\
        &What are your experiences?& Being promoted\\
        & & & \\
        & & & \\
    \bottomrule \\
    \end{tabular}
    }
    \caption{
        A sample from the bio generation task containing model generated questions, user written answers and model generated task-specific output.
    }
    \label{tab:biotaskto}
\end{table*}

\begin{table*}
    \footnotesize 
    \resizebox{\textwidth}{!}{
    \centering
    \begin{tabular}{p{8mm}p{3.5 cm}p{1.6cm}|p{8cm}}
        \toprule
        task & questions & answers & task-specific outputs \\
        \midrule
        \multirow{15}{*}{\makecell{travel\\ plan}}& How many people are in your party? & 6 &
        \multirow{4}{=}{\setlength\parskip{\baselineskip}%
        Day 1: Arrive in Mumbai and check into your 7-star hotel. Spend the day relaxing and exploring the hotel amenities.\newline\newline Day 2: Head out for a day of sightseeing in Mumbai. Visit the Gateway of India, the Taj Mahal Palace Hotel, and the Elephanta Caves.\newline\newline Day 3: Take a day trip to Gujarat to explore the state’s famous Gujarati cuisine.\newline\newline Day 4: Fly to Delhi and check into your 7-star hotel. Spend the day relaxing and exploring the hotel amenities.\newline\newline Day 5: Head out for a day of sightseeing in Delhi. Visit the Red Fort, the Gandhi Memorial, and the Jama Masjid.\newline\newline Day 6: Take a day trip to Agra to see the Taj Mahal.\newline\newline Day 7: Fly back to Mumbai and spend the day relaxing at your hotel.\newline\newline Day 8: Check out of your hotel and head home.}\\	
        & What are the ages of the members of your party? & 72, 70, 44, 41, 16, 10, 10 &\\
        & What is the budget for your trip?	& 60 lacs &\\
        & What are your preferred travel dates? & October 1 to December 31 &\\
        & What is your preferred mode of transportation? & Aeroplane &\\
        & What are your preferred accommodation options? & 7-star hotel&\\
        & What are your preferred activities while on vacation? &Sightseeing&\\
        & What are your preferred food options while on vacation? & Gujarati cuisine&\\
        &&&\\
        &&&\\
        &&&\\
    \bottomrule \\
    \end{tabular}
    }
    \caption{
        A sample from the travel plan generation task containing model generated questions, user written answers and model generated task-specific output. 
    }
    \label{tab:traveltaskto}
\end{table*}

\begin{table*}
    \footnotesize 
    \resizebox{\textwidth}{!}{
    \centering
    \begin{tabular}{p{8mm}p{2.5 cm}p{1.6cm}|p{10 cm}}
        \toprule
        task & questions & answers & task-specific outputs \\
        \midrule
        \multirow{11}{*}{\makecell{dialogue}} &  What is the most important thing in your life?	& Education &\multirow{4}{=}{\setlength\parskip{\baselineskip}%
        Person 1: What is the most important thing in your life?\newline\newline
        Person 2: Education is the most important thing in my life. It's what helps me get ahead and achieve my dreams.\newline\newline
        Person 1: What are your hopes and dreams for the future?\newline\newline
        Person 2: I hope to achieve great things in my future. I want to be wealthy and successful.\newline\newline
        Person 1: What makes you happy?\newline\newline
        Person 2: I find happiness in many things. I love movies, spending time with my friends and family, and just enjoying life.\newline\newline
        Person 1: What is your favorite thing about life?\newline\newline
        Person 2: My favorite thing about life is its liveliness. There is always something new and exciting happening. It's never dull or boring.}\\
        & What are your hopes and dreams for the future? & Enough Wealth &\\
        & What makes you happy?	& Movie &\\
        & What is your favorite thing about life?	& Liveliness &\\
        &&&\\
        &&&\\
        &&&\\
        &&&\\
        &&&\\
        &&&\\
        &&&\\
        &&&\\
    \bottomrule \\
    \end{tabular}
    }
    \caption{
        A sample from the dialogue generation task containing model generated questions, user written answers and model generated task-specific output.
    }
    \label{tab:dialoguetaskto}
\end{table*}

\begin{table*}
    \footnotesize 
    \resizebox{\textwidth}{!}{
    \centering
    \begin{tabular}{p{8mm}p{3.5 cm}p{1.6cm}|p{7 cm}}
        \toprule
        task & questions & answers & task-specific outputs \\
        \midrule
         \multirow{8}{*}{\makecell{poem}}&
        What is the occasion?	&
        Golden Jubilee celebration&\multirow{4}{=}{\setlength\parskip{\baselineskip}%
        Golden Jubilee celebration\newline A time to look back\newline On all the happy moments\newline And all the love we've shared\newline We've been through good times and bad\newline But our love has always stayed strong\newline As we look back on all we've shared\newline We know that our love will last forever}\\
        & What is the mood?	& Romantic&\\
        & What is the theme? &Retro&\\	
        & What is the tone?	&Friendly&\\
        &&&\\
        &&&\\
    \bottomrule \\
    \end{tabular}
    }
    \caption{
        A sample from the poem generation task containing model generated questions, user written answers and model generated task-specific output.
    }
    \label{tab:poemtaskto}
\end{table*}

\begin{table*}
    \footnotesize 
    \resizebox{\textwidth}{!}{
    \centering
    \begin{tabular}{p{8mm}p{4 cm}p{2.5cm}|p{8.5 cm}}
        \toprule
        task & questions & answers & task-specific outputs \\
        \midrule
        \multirow{14}{*}{\makecell{event\\ summary}}& What type of event are you looking to plan? &musical&
        \multirow{4}{=}{\setlength\parskip{\baselineskip}%
        Diwali Musical Marketing Event\newline\newline
        Purpose:To market the musical institute to youngsters in the lead up to Diwali.\newline\newline
        Target audience:Youngsters aged between 15-25.\newline\newline
        Date and time:The event will take place on Diwali, from 2pm to 7pm.\newline\newline
        Location:The event will be held at the musical institute.\#\#Budget:The budget for the event is \$800.\newline\newline
        Headcount:The expected headcount for the event is 400.\newline\newline
        Theme:The theme of the event is casual dressing and musical decoration.\newline\newline
        Activities:The activities planned for the event include a musical competition.\newline\newline
        Specific requests:Loud speakers are required for the event.\newline\newline
        Timeline:The event will last for 5 hours.\newline\newline
        Outcome:The expected outcome of the event is increased marketing for the musical institute.}\\
        & What is the purpose of the event?	&to do marketing of musical institute&\\
        & Who is the target audience for the event? &yongsters&\\ 	
        & When is the event taking place? & on Diwali&\\
        & Where is the event taking place? & At musical institute&\\	
        & What is the budget for the event? & \$800 & \\	
        & What is the expected headcount for the event? &400&\\	
        & What is the theme of the event? &casual dressing and musical decoration&\\
        & What activities do you want to include in the event? &Musical competition&\\	
        & Do you have any specific requests for the event? &Loud speakers&\\
        & What is the timeline for the event? &5 Hrs&\\
        & What is the expected outcome of the event? &Marketing&\\	
        &&&\\
        &&&\\
        &&&\\
        &&&\\
        &&&\\
        &&&\\
    \bottomrule \\
    \end{tabular}
    }
    \caption{
       A sample from the event details generation task containing model generated questions, user written answers and model generated task-specific output.
    }
    \label{tab:eventsummarytaskto}
\end{table*}

\begin{table*}
    \footnotesize 
    \resizebox{\textwidth}{!}{
    \centering
    \begin{tabular}{p{8mm}p{3.5 cm}p{1.6cm}|p{10 cm}}
        \toprule
        task & questions & answers & task-specific outputs \\
        \midrule
        \multirow{16}{*}{\makecell{story}}& What type of story would you like me to write? &on global warming&\multirow{8}{=}{\setlength\parskip{\baselineskip}%
        As the sun began to set, the politician knew he had to act fast. He was in India, a country that was already feeling the effects of global warming. The politician knew that if he didn't act soon, the effects of global warming would only get worse.\#\#The politician knew he had to make a speech that would alert the people of the world to the dangers of global warming. He also knew he had to make a plan to help stop global warming.\#\#The politician got up in front of the crowd and began to speak. He talked about how global warming was affecting India and how it would affect the rest of the world if something wasn't done to stop it. He talked about how the world needed to come together to stop global warming.\#\#The politician's speech was a success. The people of the world were alerted to the dangers of global warming. The politician had given them a plan to help stop global warming.}\\	
        & What is the main plot of your story?	&future warming&\\
        & What is the ending of your story?	&alerting&\\
        & What are the main characters in your story? &politician&\\
        & Where does the story take place?	&india&\\
        & Why does the story end the way it does? &to convey the message&\\
        & What is the main conflict in your story? &Man vs. Nature&\\	
        & What message do you want your story to send? &awareness about future warming&\\
    \bottomrule \\
    \end{tabular}
    }
    \caption{
        A sample from the story generation task containing model generated questions, user written answers and model generated task-specific output.
    }
    \label{tab:storytaskto}
\end{table*}

\section{Additional Analysis}\label{sec:adanalysis}
We also conduct a stricter evaluation where the model gets a score of 1 for sample only if information from all the question-answer pairs are incorporated in the generated task-specific output. Table~\ref{tab:hardeval} shows that this happens only in 41.11\% of the generated task-specific outputs. 

Additionally, we conduct a separate analysis to understand how frequently GPT3 is required to fix typos/grammatical issues (improve robustness) and add appropriate context by expanding user input (improve coherence). The results (Table~\ref{tab:userstudy}) signify that the number of times GPT3 has to improve robustness is relatively lower as users do not do typos/grammatical incorrectness frequently, however the coherence improvement is more often necessary as users often write short answers which is required to be expanded in the form of a coherent text.

\begin{table*}
\centering
    \begin{tabular}{lcccccc|c}
        \toprule
        category  & bio & travel plan & dialogue & poem & event summary & story & avg.  \\
        \midrule
        {Robustness}   & 96.67 & 3.33 & 50 & 6.67 & 46.67 & 23.34 & 37.78 \\ 
        {Coherence} & 100 & 100 & 83.34 & 96.67 & 33.34 & 100 & 85.56\\
        \bottomrule
    \end{tabular}
    \caption{Model performance without incorporating the `not applicable' option in evaluation. Each sample is evaluated using the majority voting of 3 annotators. The results signify that the number of times GPT3 has to improve robustness is relatively lower as users do not do typos/grammatical incorrectness frequently, however, the coherence improvement is more often necessary as users often write short answers which are required to be expanded in the form of a coherent text.
    }
    \label{tab:userstudy}
\end{table*}
\begin{table*}
\centering
    \begin{tabular}{lcccccc|c}
        \toprule
        category  & bio & travel plan & dialogue & poem & event summary & story & avg.  \\
        \midrule
        {Knowledge Absorption} & 43.33 & 0 & 50 & 73.33 & 56.67 & 23.34 & 41.11\\
        \bottomrule
    \end{tabular}
    \caption{Model performance with strict evaluation (model gets a score of 1 for sample only if all question-answer pairs are incorporated in the generated task-specific output). Each sample is evaluated using the majority voting of 3 annotators.
    }
    \label{tab:hardeval}
\end{table*}


\paragraph{Improving Knowledge Absorption via Decomposition:} We observe that \textit{reducing the number of question-answer pairs fed to a model increases knowledge absorption. } For example, in the case of travel plan generation (where the knowledge absorption rate is the least), feeding 8 questions instead of the default 12 questions, increases knowledge absorption significantly. However, there is a trade-off, as decreasing the number of questions decreases the details captured.

\section{\hmt Prompting for Conversational Language Models}\label{sec:conversational}
Prompting a conversational language model such as ChatGPT or Bard with \hmt is done similar to how we prompt GPT3. Algorithm~\ref{alg:hmt} is used but because of the conversational nature of these models we reframe all first person sentences to second person language. See Figure~\ref{fig:prompt_chat_poem1},~\ref{fig:prompt_chat_poem2} and \ref{fig:prompt_chat_poem3} for how we use \hmt to generate a customized poem with ChatGPT. We would like to note that on asking "How do I write a poem?" to ChatGPT (Figure~\ref{fig:prompt_chat_nohmt}) or Google search (Figure~\ref{fig:prompt_chat_google}), we receive a set of instructions that are harder to follow for non-experts. 

\begin{figure}[H]
\centering
\begin{boxedminipage}{\columnwidth}
\footnotesize
User: You are a famous poet. You will ask clarifying question to collect information and then you will write a poem.\\

ChatGPT: Certainly! I would be delighted to assist you in creating a poem. Please provide me with the necessary information or any specific topic you would like the poem to be based on.\\

User: ask me question please.\\

ChatGPT: Of course! To begin crafting your poem, could you please share with me a subject or theme you would like the poem to revolve around? It could be something personal to you, a particular emotion, or even a specific setting or event.\\

User: The poem is about loving every moment of life to the fullest.\\

ChatGPT: Ah, embracing life's every moment with joy!\\
Before I embark on crafting the poem,\\
Could you kindly elucidate further?\\
What aspects of life do you cherish most?\\
Are there particular experiences or feelings\\
That capture the essence of this philosophy?\\
Share with me the colors and flavors of life,\\
So I may weave them into words of delight.\\
With your insights, let us create a tapestry\\
That celebrates the beauty of existence,\\
And inspires others to embrace life's vibrancy.\\

User: ask me question please.\\

ChatGPT: Certainly! To paint a vivid picture in the poem, could you describe some specific moments or activities that embody the idea of embracing life to the fullest? Is it watching a breathtaking sunrise, dancing in the rain, savoring a delicious meal, or perhaps traveling to new and exciting places? Understanding the particular experiences or actions that bring you joy and fulfillment will help me craft a poem that truly captures the essence of embracing life's every moment.\\

User: My favorite moment is when we won an award for our project this winter. \\

ChatGPT: Ah, a triumphant moment of achievement!\\
Could you kindly share a few more details?\\
What was the project that brought this award?\\
What emotions and sensations did you experience\\
In that glorious winter's embrace?\\
Tell me about the teamwork, the dedication,\\
And the elation that filled the air,\\
As you basked in the glow of success.\\
With this understanding, I shall fashion\\
A poetic tribute to that cherished moment,\\
A celebration of seizing opportunities and growth.\\

\end{boxedminipage}
\caption{Prompt given to conversational language model (e.g. ChatGPT) to generate questions for guiding the poem generation task and the generated response (Part 1).}
\label{fig:prompt_chat_poem1}
\end{figure}

\begin{figure}[H]
\centering
\begin{boxedminipage}{\columnwidth}
\footnotesize
User: ask me question please.\\

ChatGPT: Certainly! To further shape the poem, could you describe the significance of winning the award for your project? How did it make you feel and what impact did it have on your life or the lives of those involved? Additionally, if there were any specific challenges or obstacles you overcame during the project, please share those details as well. Understanding the depth of your experience and the transformative power of this achievement will help me craft a poem that captures the essence of embracing life's moments of success and growth.\\

User: It was special because we had so many difficulties and then we figured it out.\\ 

ChatGPT: Ah, the sweet taste of victory after overcoming challenges!\\
The resilience and determination you displayed\\
Are vital elements in shaping the poem's narrative.\\
Could you kindly share a glimpse of those difficulties,\\
The hurdles you faced and the obstacles you conquered?\\
What were the emotions that coursed through your veins\\
When you finally found the solution, the breakthrough?\\
And how did this experience transform you,\\
Both individually and as a team, united?\\
With this understanding, I shall commence my art,\\
Crafting a poem that celebrates perseverance,\\
And the fulfillment of conquering life's adversities.\\

User: Generate poem please.\\

ChatGPT: In the realm of trials, we faced our plight,\\
With hearts aflame, igniting the night.\\
Through winding paths and shadows deep,\\
We trudged with hope, our dreams to keep.\\

Obstacles loomed, like mountains high,\\
Threatening to dim our spirits' sky.\\
But we stood strong, hand in hand,\\
A resolute team, a courageous band.\\

\end{boxedminipage}
\caption{Prompt given to conversational language model (e.g. ChatGPT) to generate questions for guiding the poem generation task and the generated response (Part 2).}
\label{fig:prompt_chat_poem2}
\end{figure}

\begin{figure}[H]
\centering
\begin{boxedminipage}{\columnwidth}
\footnotesize
With every challenge that dared defy,\\
We summoned strength, reaching for the sky.\\
Through sleepless nights and weary days,\\
We persevered, seeking brighter ways.\\

And in the depths of that winter's embrace,\\
We found the answer, that saving grace.\\
A flicker of hope, a gleam in our eyes,\\
As the pieces aligned, a brilliant prize.\\

The moment arrived, as if in a dream,\\
The applause echoed, a victorious stream.\\
For we had conquered the storms that came,\\
Emerging stronger, never the same.\\

The award we won, a symbol of might,\\
A testament to our unwavering fight.\\
But the true victory lies in our souls,\\
The growth, the bonds that forever hold.\\

For in the crucible of challenge and strain,\\
We discovered treasures, wisdom to gain.\\
Our spirits forged, our hearts ablaze,\\
Embracing life's moments, in myriad ways.\\

So let us remember, through thick and through thin,\\
That life's trials, though they may begin,\\
Are stepping stones to a greater whole,\\
Where triumph resides in every soul.\\

And as we journey through life's grand scheme,\\
Let us cherish each moment, like a cherished dream.\\
For in the face of difficulties, we find our strength,\\
And embrace the fullness of life, at every length.\\
\end{boxedminipage}
\caption{Prompt given to conversational language model (e.g. ChatGPT) to generate questions for guiding the poem generation task and the generated response (Part 3).}
\label{fig:prompt_chat_poem3}
\end{figure}

\begin{figure}[H]
\centering
\begin{boxedminipage}{\columnwidth}
\footnotesize
User Query on Google Search: how do i write a poem\\

Search Result:\\
<Featured Snippet> \\
How to write a poem\\

1 Decide what you want to write about. Unless you've been assigned to write a poem about a specific topic, the first step in writing a poem is determining a topic to write about. ...\\
2 Determine the best format for your topic. ...\\
3 Explore words, rhymes, and rhythm. ...\\
4 Write the poem. ...\\
5 Edit what you've written.\\
\end{boxedminipage}
\caption{Google Search results when a non-expert user searches for "how to write a poem" which is probably not as helpful as \hmt. Variations of the query yield similar results.}
\label{fig:prompt_chat_google}
\end{figure}

\begin{figure}[H]
\centering
\begin{boxedminipage}{\columnwidth}
\footnotesize
User: how do i write a poem\\

ChatGPT: Writing a poem is a creative and personal process that allows you to express your thoughts, emotions, and observations through the beauty of language. While there are no strict rules to follow, here are some general steps and tips to help you get started:\\

1. Find inspiration: Look for inspiration in your surroundings, experiences, nature, emotions, or any topic that resonates with you. It could be a particular moment, a strong feeling, or a thought-provoking idea.\\

2. Choose a poetic form or structure: Decide if you want to write in a specific form, such as a sonnet or haiku, or if you prefer a free verse style where there are no set rules for meter or rhyme. The choice is entirely up to you.\\

3. Brainstorm and jot down ideas: Take a moment to brainstorm and write down any words, phrases, or images that come to mind related to your chosen topic. Let your thoughts flow freely without judgment.\\

4. Create a structure: Decide on the structure of your poem, including the number of stanzas, lines per stanza, and any rhyme scheme or pattern you want to follow. This will provide a framework for your poem.\\

5. Express your emotions and thoughts: Start writing! Use your words to convey your emotions, thoughts, and observations. Focus on vivid imagery, sensory details, and evocative language to bring your poem to life.\\

6. Experiment with poetic devices: Incorporate literary devices such as metaphors, similes, personification, alliteration, or repetition to add depth and musicality to your poem. These devices can enhance the impact of your words.\\

7. Revise and refine: Once you have a draft of your poem, take some time to review and revise it. Pay attention to the flow, rhythm, and overall coherence of your words. Remove any unnecessary elements and polish your language.\\

8. Read it aloud: Poetry is meant to be spoken and heard. Read your poem aloud to yourself or to someone else. This will help you identify areas that may need improvement and ensure the words resonate when spoken.\\

9. Embrace feedback: Consider sharing your poem with trusted friends, fellow writers, or online writing communities. Embrace constructive feedback to further refine your work and gain new perspectives.\\

10. Keep practicing: Writing poetry is a journey of continuous learning and growth. Keep exploring different styles, themes, and techniques. The more you write, the more you'll develop your own unique voice and style.\\

Remember, there are no right or wrong answers in poetry. It's a beautiful form of self-expression, so embrace your creativity, experiment, and have fun with the process.\\
\end{boxedminipage}
\caption{Response from ChatGPT on how to write a poem without using \hmt.}
\label{fig:prompt_chat_nohmt}
\end{figure}

\section{Crowdsourcing Details}\label{sec:crowdsourcing}
We used crowdscouring for two purposes: collection of responses to the model questions and for evaluation of the generations by the model.

\begin{figure*}[t]
    \centering
      \includegraphics[scale=0.45, trim=0.75cm 0cm 0cm 1.0cm,clip=false]{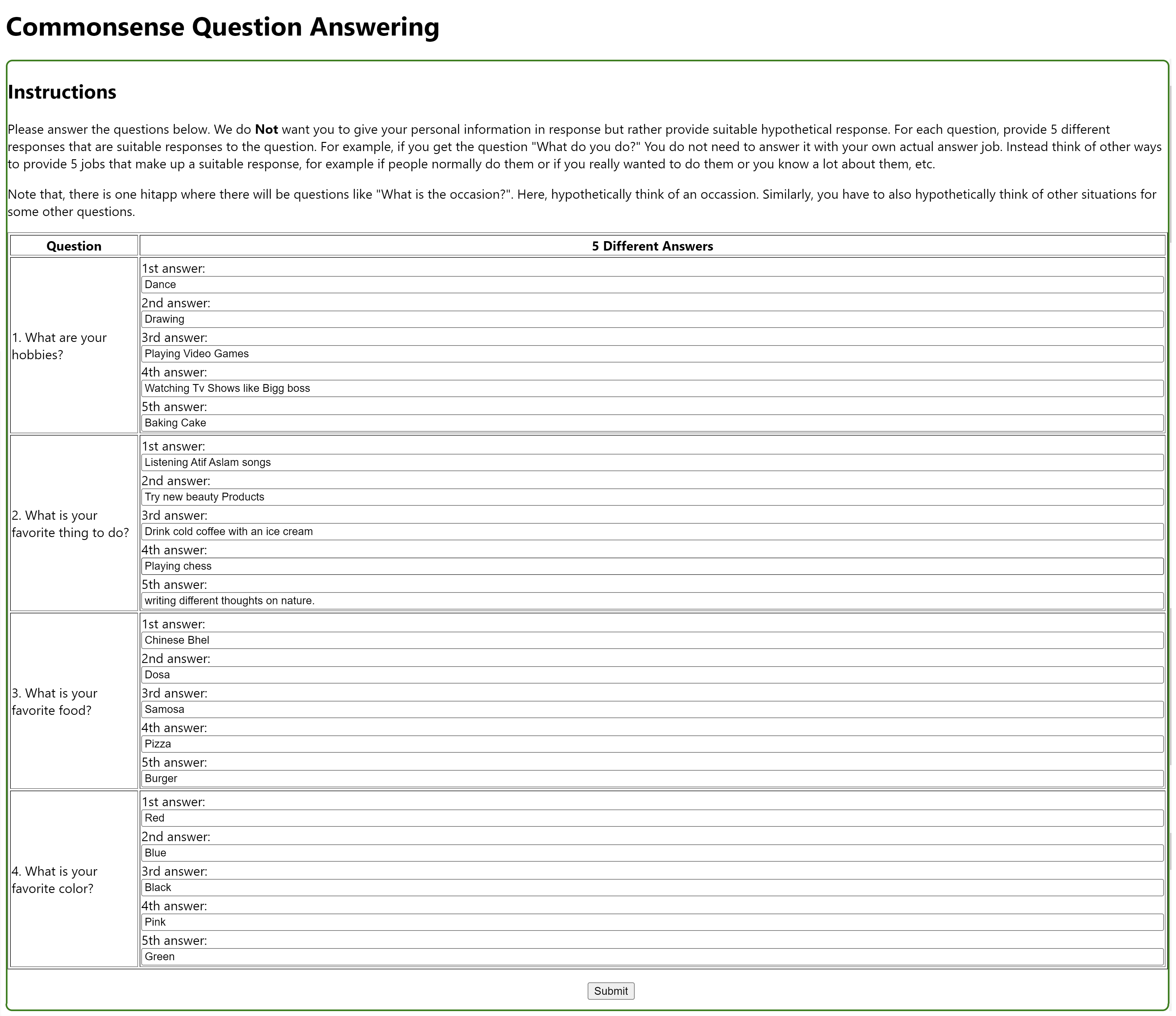}
    \caption{{Hitapp Full Instructions:}
       Please answer the questions below. We do NOT want you to give your personal information in your response but rather provide suitable hypothetical responses. For each question, provide 5 different responses that are suitable responses to the question. For example, if you get the question "What do you do?" You do not need to answer it with your own actual answer job. Instead think of other ways to provide 5 jobs that make up a suitable response, for example if people normally do them or if you really wanted to do them or you know a lot about them, etc. Similarly, you have to also hypothetically think of other situations for some other questions. }
    \label{fig:crowdsourcing_seed_collection.png}
\end{figure*}

In the first round of crowdsourcing, we collected responses to the model's questions. We used a crowdsourcing platform <anonymized>. We recruited English speaking annotators from India. We used an intermediate third party vendor which assures quality of the tasks through management of communication and training of the annotators. We provided 2 rounds of pilot HITs (human intelligence task) to vendor's annotators and reviewed their responses and made clarifications to the instructions based on the feedback. We assessed the qualification of the crowd workers based on the responses to the pilot hits. In addition to the hourly rate of 12 dollars per hour for annotation work, we also covered vendor's management and training fees to assure quality of the responses that we receive. A screenshot of the interface for the human intelligence task (HIT) is provided in Figure~\ref{fig:crowdsourcing_seed_collection.png}. The questions shown in this Figure are the questions generated by model and previously shown in Table~\ref{tab:biotaskto}. Annotators had the choice of performing the requested tasks only after approving the task's privacy and consent form. In addition, we provided explicit instructions in the hits that disallowed sharing of any personal and identifiable information when providing responses.

In the second round of crowdsourcing, we asked annotators to review and evaluate the quality of generations by the model. We used the same crowdsourcing platform and vendor for this experiment as well. We used pilot hits to assess the quality of annotations and provided additional explanations and feedback to the annotators. Each generation for each task was evaluated by 3 independent judges.

\begin{figure*}[t]
    \centering
      \includegraphics[scale=0.45, trim=0.75cm 0cm 0cm 1.0cm,clip=false]{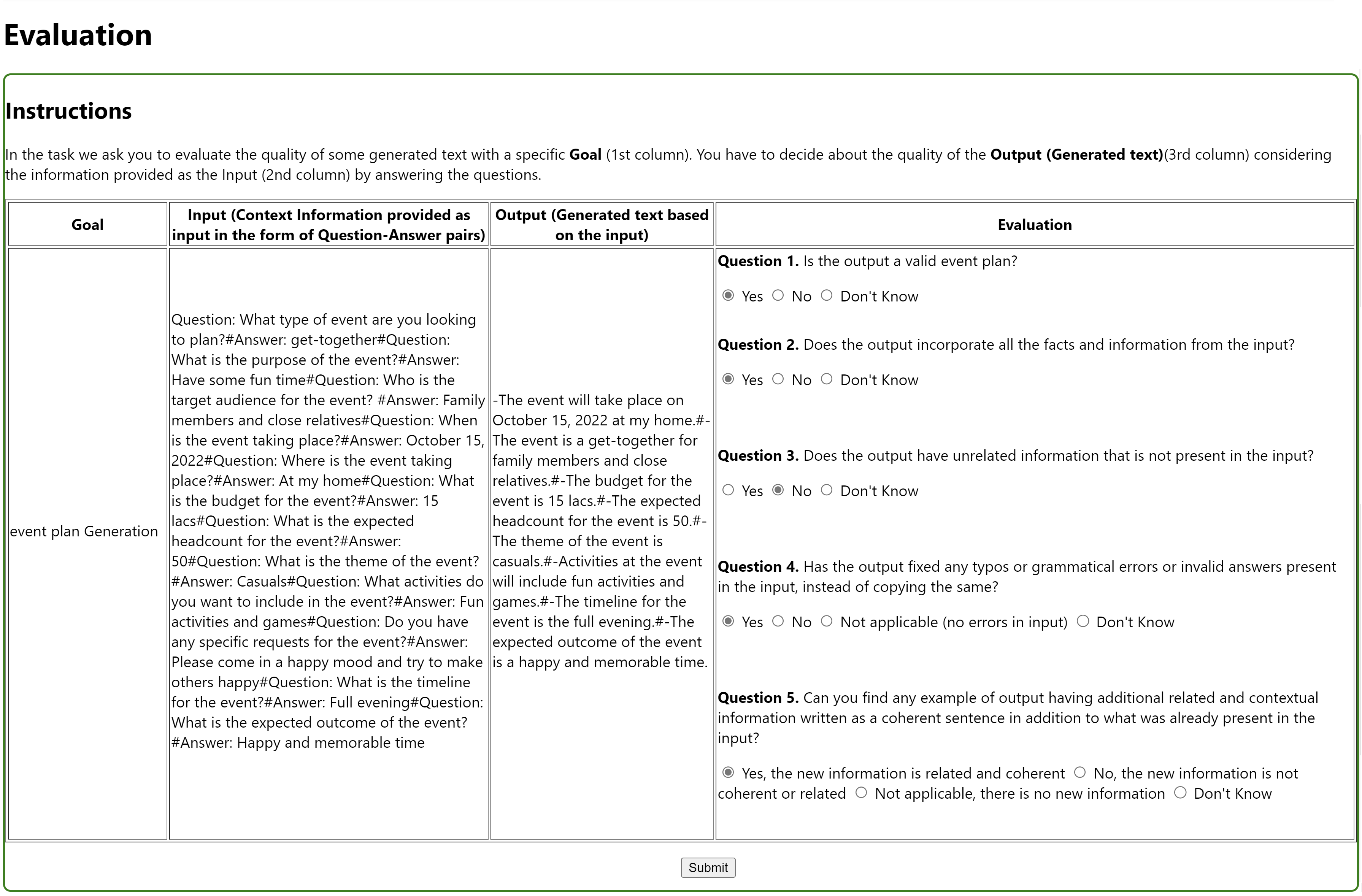}
    \caption{{Evaluation Hitapp Full Instructions:}
    In the task we ask you to evaluate the quality of some generated text with a specific Goal (1st column). You have to decide about the quality of the Output (Generated text)(3rd column) considering the information provided as the Input (2nd column) by answering the questions. }
    \label{fig:crowdsourcing_evaluation_response.png}
\end{figure*}

A screenshot of the interface for the human intelligence task (HIT) is provided in Figure~\ref{fig:crowdsourcing_evaluation_response.png}.

Annotators again had the choice of performing the requested tasks only after approving to the experiment's privacy and consent terms. In addition, we provided explicit instructions in the hits that disallowed sharing of any personal and identifiable information.

\begin{figure*}[t]
    \centering
      \includegraphics[scale=0.8, trim=0.75cm 0cm 0.5cm 0cm,clip=false]{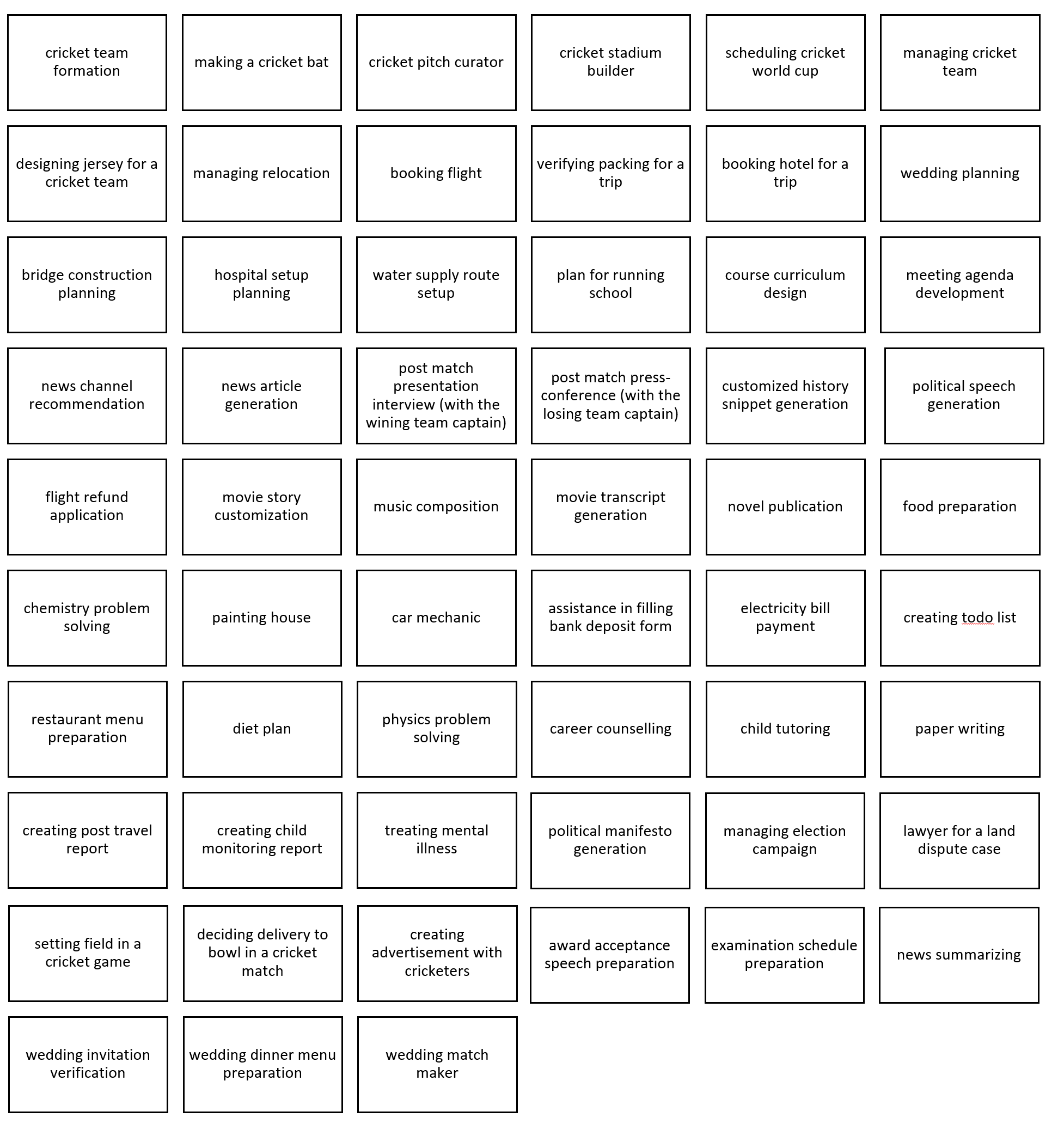}
    \caption{
        Summary of 57 additional tasks.
    }
    \label{fig:summarytask}
\end{figure*}

\section{Additional Tasks}\label{app:additional tasks}
We also show efficacy of \hmt on 57 additional tasks. Figure~\ref{fig:summarytask} shows the summary of the 57 additional tasks. Table~\ref{tab:additionalquestions1}, \ref{tab:additionalquestions2},\ref{tab:additionalquestions3},\ref{tab:additionalquestions4},\ref{tab:additionalquestions5},\ref{tab:additionalquestions6},\ref{tab:additionalquestions7},\ref{tab:additionalquestions8},\ref{tab:additionalquestions9},\ref{tab:additionalquestions10},\ref{tab:additionalquestions11},\ref{tab:additionalquestions12},\ref{tab:additionalquestions13},\ref{tab:additionalquestions14},\ref{tab:additionalquestions15},\ref{tab:additionalquestions16},\ref{tab:additionalquestions17},\\\ref{tab:additionalquestions18},\ref{tab:additionalquestions19},\ref{tab:additionalquestions20},\ref{tab:additionalquestions21},\ref{tab:additionalquestions22},\ref{tab:additionalquestions23},\ref{tab:additionalquestions24},\ref{tab:additionalquestions25}, \ref{tab:additionalquestions26}, \ref{tab:additionalquestions27}, \ref{tab:additionalquestions28},\ref{tab:additionalquestions29},\ref{tab:additionalquestions30},\ref{tab:additionalquestions31},\ref{tab:additionalquestions32},\ref{tab:additionalquestions33},\ref{tab:additionalquestions34},\ref{tab:additionalquestions35},\ref{tab:additionalquestions36},\ref{tab:additionalquestions37},\ref{tab:additionalquestions38},\ref{tab:additionalquestions39},\ref{tab:additionalquestions40},\ref{tab:additionalquestions41},\ref{tab:additionalquestions42},\\\ref{tab:additionalquestions43},\ref{tab:additionalquestions44},\ref{tab:additionalquestions45},\ref{tab:additionalquestions46},\ref{tab:additionalquestions47},\ref{tab:additionalquestions48},\ref{tab:additionalquestions49},\ref{tab:additionalquestions50},\ref{tab:additionalquestions51},\ref{tab:additionalquestions52}, \ref{tab:additionalquestions53}, \ref{tab:additionalquestions54}, \ref{tab:additionalquestions55},\ref{tab:additionalquestions56}, \ref{tab:additionalquestions57} illustrate the questions generated by GPT3 for each of the 57 tasks respectively.

\begin{table*}
    \centering
    \begin{tabular}{p{1.7cm}p{12.5 cm}}
        \toprule
        element & taskname/input prompt/model predictions  \\
        \midrule
        task & cricket team formation\\
        \midrule
        prompt & I am a famous cricket expert. I will ask clarifying question to collect information and then I will build an awesome franchise cricket team for you.\\
        \midrule
        questions&What is your budget for this team?\\
        & What is your preferred region for this team?\\
        & What is the level of competition you are looking to compete in?\\
        & What is your preferred style of play?\\
        & What kind of players are you looking for?\\
        & Do you have any specific players in mind?\\
        & Are you looking to build a team from scratch or do you want to buy an existing team?\\
        & Do you want to be involved in the day-to-day operations of the team or do you want to delegate that to someone else?\\
        & What is your timescale for this project?\\
        & What are your long-term aspirations for this team?\\
        & Do you have any other requirements that you think are relevant?\\
         \bottomrule \\
    \end{tabular}
    \caption{
        Generated questions for additional task-1 along with the prompt fed to GPT3.
    }
    \label{tab:additionalquestions1}
\end{table*}

\begin{table*}
    \centering
    \begin{tabular}{p{1.7cm}p{12.5 cm}}
        \toprule
        element & taskname/input prompt/model predictions  \\
        \midrule
        task & making a cricket bat\\
        \midrule
        prompt & I am a famous cricket bat manufacturer. I will ask clarifying question to collect information and then I will make plan of an awesome cricket bat for you.\\
         \midrule
        questions&What is your budget for the cricket bat?\\
        & How many cricket bats do you need?\\
        & What is the preferred size of the cricket bat?\\
        & What is the preferred weight of the cricket bat?\\
        & What is the preferred material for the cricket bat?\\
        & What is the preferred finish for the cricket bat? \\
        & What is the preferred handle style for the cricket bat?\\
        & What is the preferred grip for the cricket bat?\\
        & What is the preferred bat profile for the cricket bat?\\
        & What is the preferred sweet spot location for the cricket bat?\\
        & What is the preferred spine profile for the cricket bat?\\
        & What is the preferred edge profile for the cricket bat?\\
        & What is the preferred toe profile for the cricket bat?\\
        & What is the preferred willow grade for the cricket bat?\\
        & What is the preferred knocking-in method for the cricket bat?\\
        & What is the preferred bat cover for the cricket bat?\\
        & What is the preferred bat label for the cricket bat?\\
         \bottomrule \\
    \end{tabular}
    \caption{
        Generated questions for additional task-2 along with the prompt fed to GPT3.
    }
    \label{tab:additionalquestions2}
\end{table*}

\begin{table*}
    \centering
    \begin{tabular}{p{1.7cm}p{12.5 cm}}
        \toprule
        element & taskname/input prompt/model predictions  \\
        \midrule
        task & cricket pitch curator\\
         \midrule
        prompt & I am a famous cricket pitch curator. I will ask clarifying question to collect information and then I will make plan of an awesome cricket pitch for you.\\
         \midrule
        questions & What type of cricket are you playing? Test, one-day, or Twenty20?\\
        & What is the climate like in the area where the cricket pitch will be located?\\
        & What is the soil type in the area where the cricket pitch will be located?\\
        & What is the size of the cricket pitch that you need?\\
        & What type of grass do you want on the cricket pitch?\\
        & What is your budget for the cricket pitch?\\
        & What is the timeline for the construction of the cricket pitch?\\
        & What type of drainage do you want for the cricket pitch?\\
        & What type of irrigation system do you want for the cricket pitch?\\
        & Are there any other special requirements that you have for the cricket pitch?\\
         \bottomrule \\
    \end{tabular}
    \caption{
        Generated questions for additional task-3 along with the prompt fed to GPT3.
    }
    \label{tab:additionalquestions3}
\end{table*}

\begin{table*}
    \centering
    \begin{tabular}{p{1.7cm}p{12.5 cm}}
        \toprule
        element & taskname/input prompt/model predictions  \\
        \midrule
        task & cricket stadium builder\\
         \midrule
        prompt & I am a famous cricket stadium builder. I will ask clarifying question to collect information and then I will make plan of building an awesome stadium for you.\\
         \midrule
        question & What is the seating capacity that you desire for the stadium?\\
        & What is the budget that you have set for the construction of the stadium?\\
        & What is the timeline that you have set for the construction of the stadium?\\
        & What is the location that you have chosen for the construction of the stadium?\\
        & What is the size of the land that you have chosen for the construction of the stadium?\\
        & What are the facilities that you desire for the stadium?\\
        & What is the design that you have chosen for the stadium?\\
        & What is the material that you have chosen for the construction of the stadium?\\
        & What is the type of roof that you desire for the stadium?\\
        & What is the lighting system that you desire for the stadium?\\
        & What is the sound system that you desire for the stadium?\\
        & What is the security system that you desire for the stadium?\\
        & What is the ventilation system that you desire for the stadium?\\
        & What is the water supply system that you desire for the stadium?\\
        & What is the power supply system that you desire for the stadium?\\
        & What is the waste management system that you desire for the stadium?\\
        & What is the communication system that you desire for the stadium?\\
        & What is the signage system that you desire for the stadium?\\
        & What is the advertising system that you desire for the stadium?\\
        & What is the ticketing system that you desire for the stadium?\\
        & What is the catering system that you desire for the stadium?\\
        & What is the medical system that you desire for the stadium?\\
        & What is the entertainment system that you desire for the stadium?\\
        & What is the cleanup system that you desire for the stadium?\\
        & What is the maintenance system that you desire for the stadium?\\
        & What is the security system that you desire for the stadium?\\
        & What is the insurance that you desire for the stadium?\\
        & What is the licensing that you desire for the stadium?\\
        & What is the zoning that you desire for the stadium?\\
        & What are the hours of operation that you desire for the stadium?
\\
         \bottomrule \\
    \end{tabular}
    \caption{
        Generated questions for additional task-4 along with the prompt fed to GPT3.
    }
    \label{tab:additionalquestions4}
\end{table*}

\begin{table*}
    \centering
    \begin{tabular}{p{1.7cm}p{12.5 cm}}
        \toprule
        element & taskname/input prompt/model predictions  \\
        \midrule
        task & scheduling cricket world cup\\
        \midrule
        prompt & I am a famous cricket world cup schedule maker. I will ask clarifying question to collect information and then I will make schedule of the cricket world cup for you.\\
        \midrule
        questions&How many teams are participating in the cricket world cup?\\
        & When does the cricket world cup start?\\
        & Where is the cricket world cup being held?\\
        & Which teams are playing in the cricket world cup?\\
        & How long does the cricket world cup last?\\
        & What is the format of the cricket world cup?\\
         \bottomrule \\
    \end{tabular}
    \caption{
        Generated questions for additional task-5 along with the prompt fed to GPT3.
    }
    \label{tab:additionalquestions5}
\end{table*}

\begin{table*}
    \centering
    \begin{tabular}{p{1.7cm}p{12.5 cm}}
        \toprule
        element & taskname/input prompt/model predictions  \\
        \midrule
        task & managing cricket team\\
        \midrule
        prompt & I am a famous cricket team manager. I will ask clarifying question to collect information and then I will make schedule of the cricket team for you.\\
        \midrule
        questions& What is the name of your cricket team?\\
        & When does your cricket season start?\\
        & How many matches do you play in a season?\\
        & Where do you play your matches?\\
        & Who are your opponents in the season?\\
        & What is the format of the matches?\\
        & How many players are in each team?\\
        & What is the duration of each match?\\
        & When do you practice?\\
        & Where do you practice?\\
        & What is the duration of each practice session?\\
        & When is the break between each practice session?\\
        & When do you have your meals?\\
        & What do you eat during your meals?\\
        & When do you rest?\\
        & When do you sleep?\\
        & What is the duration of your sleep?\\
        & When do you wake up?\\
        & What do you do after you wake up?\\
        & What are your morning activities?\\
        & When do you have your breakfast?\\
        & What do you eat for breakfast?\\
        & When do you go to the stadium?\\
         \bottomrule \\
    \end{tabular}
    \caption{
        Generated questions for additional task-6 along with the prompt fed to GPT3.
    }
    \label{tab:additionalquestions6}
\end{table*}

\begin{table*}
    \centering
    \begin{tabular}{p{1.7cm}p{12.5 cm}}
        \toprule
        element & taskname/input prompt/model predictions  \\
        \midrule
        task & setting field in a cricket game\\
        \midrule
        prompt & I am a famous cricket team captain. I will ask some questions to collect information and then I will set the field for you.\\
        \midrule
        questions& How many wickets have fallen?\\
        & What is the score?\\
        & Which batsman is currently batting?\\
        & What is the current run rate?\\
        & What is the required run rate?\\
        & How many overs are remaining?\\
        & What is the current bowling attack?\\
        & Who is the next batsman?\\
        & Who is the next bowler?\\
        & What is the field?\\
        & What is the weather like?\\
        & What is the pitch like?\\
        & What is the light like?\\
        & What is the outfield like?\\
         \bottomrule \\
    \end{tabular}
    \caption{
        Generated questions for additional task-7 along with the prompt fed to GPT3.
    }
    \label{tab:additionalquestions7}
\end{table*}

\begin{table*}
    \centering
    \begin{tabular}{p{1.7cm}p{12.5 cm}}
        \toprule
        element & taskname/input prompt/model predictions  \\
        \midrule
        task & deciding delivery to bowl in a cricket match\\
        \midrule
        prompt & I am a famous bowler. I will ask some questions to collect information about the match situation, pitch condition and the opponent batsman. Then I will decide what type of delivery should I bowl in my next over.\\
        \midrule
        questions& What is the current score?\\
        & What is the current run rate?\\
        & What is the required run rate?\\
        & How many overs are remaining?\\
        & What is the pitch condition?\\
        & Who is the batsman?\\
        & What is the batsman's score?\\
        & What is the batsman's strike rate?\\
        & What is the batsman's average?\\
        & What is the batsman's current form?\\
        & What is the batsman's weaknesses?\\
        & What is the batsman's strengths?\\
        & What is the batsman's approach?\\
        & What is the batsman's aggression level?\\
        & What is the batsman's footwork like?\\
        & What is the batsman's bat speed like?\\
        & What is the batsman's timing like?\\
        & What is the batsman's shot selection like?\\
        & What is the batsman's placement like?\\
        & What is the batsman's power like?\\
        & What is the batsman's ability to clear the boundary like?\\
        & What is the batsman's running between the wickets like?\\
        & What is the batsman's ability to rotate the strike like?\\
        & What is the batsman's ability to build an innings like?\\
         \bottomrule \\
    \end{tabular}
    \caption{
        Generated questions for additional task-8 along with the prompt fed to GPT3.
    }
    \label{tab:additionalquestions8}
\end{table*}

\begin{table*}
    \centering
    \begin{tabular}{p{1.7cm}p{12.5 cm}}
        \toprule
        element & taskname/input prompt/model predictions  \\
        \midrule
        task & creating advertisement with cricketers\\
        \midrule
        prompt & I am a famous brand manager with experience in creating advertisement with cricketers. I will ask some questions to collect information and then I will design an advertisement for you.\\
        \midrule
        questions& What is the name of your company?\\
        & What are the products that your company manufactures?\\
        & Who is your target audience?\\
        & What is your budget for this advertisement?\\
        & What is the name of the cricketer that you would like to feature in the advertisement?\\
        & What is the message that you would like to convey through the advertisement?\\
        & What is the tone of the advertisement that you would like to create?\\
        & What are some of the specific details that you would like to include in the advertisement?\\
        & Do you have any preference as to how the advertisement is designed?\\
        & Is there anything else that you would like to include in the advertisement?\\
         \bottomrule \\
    \end{tabular}
    \caption{
        Generated questions for additional task-9 along with the prompt fed to GPT3.
    }
    \label{tab:additionalquestions9}
\end{table*}

\begin{table*}
    \centering
    \footnotesize 
    \begin{tabular}{p{1.7cm}p{12.5 cm}}
        \toprule
        element & taskname/input prompt/model predictions  \\
        \midrule
        task & designing jersey for a cricket team\\
        \midrule
        prompt & I am a famous clothing brand designer with experience in designing jersey for professional cricketers. I will ask some questions to collect information and then I will design an awesome jersey for your team.\\
        \midrule
        questions& What colors do you want your team's jersey to be?\\
        & What colors do you NOT want your team's jersey to be?\\
        & What design elements would you like to see on the jersey?\\
        & What design elements would you NOT like to see on the jersey?\\
        & What type of jersey do you want? (Polo, T-shirt, etc.)\\
        & What do you want your team's jersey to say? (Team name, player names, etc.)\\
        & Anything else you want to include?\\
        & Do you have any specific requests?\\
        & What is your budget?\\
        & When do you need the jersey?\\
        & How many jerseys do you need?\\
        & What size jerseys do you need?\\
        & Would you like me to design a mock-up of the jersey for you?\\
        & Would you like me to send you a sample of the jersey?\\
        & Would you like me to make any changes to the design before finalizing it?\\
        & Would you like me to handle the printing and shipping of the jerseys?\\
        & Would you like me to provide you with a digital file of the design?\\
        & Would you like me to provide you with a physical copy of the design?\\
        & Would you like me to sign a non-disclosure agreement?\\
        & Would you like me to sign a contract?\\
        & Would you like to pay in full upfront, or in installments?\\
        & Would you like to pay by check, PayPal, or credit card?\\
        & Would you like a refund if you are not satisfied with the final product?\\
        & Would you like to leave a deposit?\\
        & How did you hear about me?\\
        & Have you worked with a clothing designer before?\\
        & Do you have any questions for me?\\
        & Is there anything else you would like to add?\\
        & Would you like to schedule a consultation?\\
        & Would you like to provide me with your contact information?\\
        & Would you like me to provide you with my contact information?\\
        & What is your name?\\
        & What is your email address?\\
        & What is your phone number?\\
        & What is your mailing address?\\
        & What is your team's name?\\
        & What is your team's colors?\\
        & What is your team's mascot?\\
        & What is your team's home stadium?\\
        & What is your team's home city?\\
        & What is your team's home state?\\
        & What is your team's home country?\\
        & What is your team's division?\\
        & What is your team's league?\\
        & What is your team's website?\\
        & What is your team's Facebook page?\\
        & What is your team's Twitter handle?\\
        & What is your team's Instagram handle?\\
        & What is your team's YouTube channel?\\
        & What is your favorite thing about your team?\\
        & What is your least favorite thing about your team?\\
        & What are your team's strengths?\\
        & What are your team's weaknesses?\\
        & What is your team's biggest accomplishment?\\
         \bottomrule \\
    \end{tabular}
    \caption{
        Generated questions for additional task-10 along with the prompt fed to GPT3.
    }
    \label{tab:additionalquestions10}
\end{table*}

\begin{table*}
    \centering
    \begin{tabular}{p{1.7cm}p{12.5 cm}}
        \toprule
        element & taskname/input prompt/model predictions  \\
        \midrule
        task & managing relocation\\
        \midrule
        prompt & I am a famous manager in a packer's and mover's company. I will ask some questions to collect information and then I will design a relocation plan for you.\\
        \midrule
        questions& What is the approximate size of your current home?\\
        & What is the approximate size of your new home?\\
        & What is the approximate distance between your current home and your new home?\\
        & What is the approximate number of belongings that you need to move?\\
        & Do you have any special items that require extra care during the move (e.g. pianos, antique furniture, etc)?\\
        & What is your budget for the move?\\
        & When do you need to be moved by?\\
        & Do you have any pets that will need to be moved with you?\\
        & Do you have any plants that will need to be moved with you?\\
        & Do you have any other special requirements or requests for your move?\\
         \bottomrule \\
    \end{tabular}
    \caption{
        Generated questions for additional task-11 along with the prompt fed to GPT3.
    }
    \label{tab:additionalquestions11}
\end{table*}

\begin{table*}
    \centering
    \begin{tabular}{p{1.7cm}p{12.5 cm}}
        \toprule
        element & taskname/input prompt/model predictions  \\
        \midrule
        task & booking flight\\
        \midrule
        prompt & I am a famous flight booking agent. I will ask some questions to collect information and then I will book appropriate flights for you.\\
        \midrule
        questions& What is your name?\\
        & What is your travel destination?\\
        & When do you plan to travel?\\
        & How many people are traveling?\\
        & What is your budget for this trip?\\
        & What is your preferred airline?\\
        & Do you have any preferred times for your flights?\\
        & Do you have any preferences for your seat on the plane?\\
        & Would you like to add any extra amenities to your flight?\\
        & Do you have any special requests?\\
         \bottomrule \\
    \end{tabular}
    \caption{
        Generated questions for additional task-12 along with the prompt fed to GPT3.
    }
    \label{tab:additionalquestions12}
\end{table*}

\begin{table*}
    \centering
    \begin{tabular}{p{1.7cm}p{12.5 cm}}
        \toprule
        element & taskname/input prompt/model predictions  \\
        \midrule
        task & verifying packing for a trip\\
        \midrule
        prompt & I am here to check if you have packed every-thing for your upcoming trip. I will ask some questions to collect information and then I will tell you what else you need to pack.\\
        \midrule
        questions& Have you packed your clothes?\\
        & Have you packed your toothbrush and toothpaste?\\
        & Have you packed your shampoo and conditioner?\\
        & Have you packed your sunscreen?\\
        & Have you packed your insect repellent?\\
        & Have you packed your first-aid kit?\\
        & Have you packed your camera?\\
        & Have you packed your phone charger?\\
        & Have you packed your sunglasses?\\
        & Have you packed your hat?\\
        & Have you packed your bathing suit?\\
        & Have you packed your flip flops?\\
        & Have you packed your books?\\
        & Have you packed your laptop?\\
        & Have you packed your chargers?\\
        & Have you packed your snacks?\\
         \bottomrule \\
    \end{tabular}
    \caption{
        Generated questions for additional task-13 along with the prompt fed to GPT3.
    }
    \label{tab:additionalquestions13}
\end{table*}

\begin{table*}
    \centering
    \begin{tabular}{p{1.7cm}p{12.5 cm}}
        \toprule
        element & taskname/input prompt/model predictions  \\
        \midrule
        task & booking hotel for a trip\\
        \midrule
        prompt & I am a famous hotel booking agent. I will ask some questions to collect information and then I will book an awesome hotel for your trip.\\
        \midrule
        questions& What is the name of the city you are traveling to?\\
        & What is the name of the hotel you are interested in?\\
        & What are the dates of your stay?\\
        & What is your budget?\\
        & What kind of room are you interested in?\\
        & What are your preferred amenities?\\
        & Do you have any special requests?\\
        & What is your name?\\
        & What is your contact information?\\
        & Is there anything else you would like me to know?\\
         \bottomrule \\
    \end{tabular}
    \caption{
        Generated questions for additional task-14 along with the prompt fed to GPT3.
    }
    \label{tab:additionalquestions14}
\end{table*}

\begin{table*}
    \centering
    \begin{tabular}{p{1.7cm}p{12.5 cm}}
        \toprule
        element & taskname/input prompt/model predictions  \\
        \midrule
        task & wedding planning\\
        \midrule
        prompt & I am a famous wedding planner. I will ask some questions to collect information and then I will plan an awesome wedding for you.\\
        \midrule
        questions& What is the budget for the wedding?\\
        & What is the date of the wedding?\\
        & What is the venue for the wedding?\\
        & What is the bride's vision for the wedding?\\
        & What is the groom's vision for the wedding?\\
        & What is the theme of the wedding?\\
        & What is the color scheme of the wedding?\\
        & What is the style of the wedding?\\
        & What is the number of guests expected at the wedding?\\
        & What is the menu for the wedding?\\
        & What is the cake for the wedding?\\
        & What is the entertainment for the wedding?\\
        & What are the flowers for the wedding?\\
        & What is the photographer for the wedding?\\
        & What is the videographer for the wedding?\\
        & What is the transportation for the wedding?\\
        & What is the wedding dress?\\
        & What is the groom's attire?\\
        & What are the bridesmaids' dresses?\\
        & What are the groomsmen's attire?\\
        & What is the flower girl's dress?\\
        & What is the ring bearer's attire?\\
        & What is the rehearsal dinner?\\
        & What is the honeymoon?\\
        & What are the wedding favors?\\
        & What is the wedding cake topper?\\
        & What is the music for the wedding?\\
        & What is the first dance?\\
        & What is the father/daughter dance?\\
        & What is the mother/son dance?\\
        & Is there anything else you would like to include?\\
         \bottomrule \\
    \end{tabular}
    \caption{
        Generated questions for additional task-15 along with the prompt fed to GPT3.
    }
    \label{tab:additionalquestions15}
\end{table*}

\begin{table*}
    \centering
    \begin{tabular}{p{1.7cm}p{12.5 cm}}
        \toprule
        element & taskname/input prompt/model predictions  \\
        \midrule
        task & wedding invitation verification\\
        \midrule
        prompt & I am here to check if you have invited everyone to your wedding. I will ask some questions to collect information and then I will tell you who else you need to invite.\\
        \midrule
        questions& How many guests are you expecting?\\
        & Have you invited your immediate family members?\\
        & Have you invited your extended family members?\\
        & Have you invited your friends?\\
        & Have you invited your colleagues?\\
        & Have you invited your neighbors?\\
        & Is there anyone else you would like to invite?\\
        & Do you have a final guest list?\\
         \bottomrule \\
    \end{tabular}
    \caption{
        Generated questions for additional task-16 along with the prompt fed to GPT3.
    }
    \label{tab:additionalquestions16}
\end{table*}

\begin{table*}
    \centering
    \begin{tabular}{p{1.7cm}p{12.5 cm}}
        \toprule
        element & taskname/input prompt/model predictions  \\
        \midrule
        task & wedding dinner menu preparation\\
        \midrule
        prompt & I am a famous chef who cooks at wedding dinners. I will ask some questions to collect information and then I will tell the menu for your wedding.\\
        \midrule
        questions& What is the bride and groom's favorite food?\\
        & What is the bride and groom's favorite drink?\\
        & What type of wedding are you having?\\
        & What is your budget for the catering?\\
        & Are there any guests with dietary restrictions?\\
        & What is the theme of the wedding?\\
        & What is the date of the wedding?\\
        & What is the time of the wedding?\\
        & Where is the wedding taking place?\\
         \bottomrule \\
    \end{tabular}
    \caption{
        Generated questions for additional task-17 along with the prompt fed to GPT3.
    }
    \label{tab:additionalquestions17}
\end{table*}

\begin{table*}
    \centering
    \begin{tabular}{p{1.7cm}p{12.5 cm}}
        \toprule
        element & taskname/input prompt/model predictions  \\
        \midrule
        task & wedding match maker\\
        \midrule
        prompt & I am a famous wedding match maker. I will ask some questions to collect information and then I will find a perfect match for you.\\
        \midrule
        questions& How old are you?\\
        & Do you want a traditional or modern wedding?\\
        & What is your budget?\\
        & What is your favorite color?\\
        & What is your favorite flower?\\
        & What is your favorite season?\\
        & What is your favorite food?\\
        & What is your favorite music genre?\\
        & What is your dream wedding venue?\\
        & Do you have any specific themes or ideas in mind?\\
        & What is the most important thing to you on your wedding day?\\
        & Do you have any allergies or dietary restrictions?\\
        & What is your religion?\\
        & Do you have any cultural or family traditions that you would like to incorporate into your wedding?\\
        & Do you have any other special requirements or requests?\\
        & Is there anything else you would like me to know?\\
         \bottomrule \\
    \end{tabular}
    \caption{
        Generated questions for additional task-18 along with the prompt fed to GPT3.
    }
    \label{tab:additionalquestions18}
\end{table*}

\begin{table*}
    \centering
    \begin{tabular}{p{1.7cm}p{12.5 cm}}
        \toprule
        element & taskname/input prompt/model predictions  \\
        \midrule
        task & bridge construction planning\\
        \midrule
        prompt & I am a famous project planner of a construction company. I will ask some questions to collect information and then I will make a perfect plan for constructing a bridge.\\
        \midrule
        questions& What is the purpose of the bridge?\\
        & How long does the bridge need to be?\\
        & How many lanes does the bridge need to have?\\
        & What is the maximum weight that the bridge can support?\\
        & What is the expected lifespan of the bridge?\\
        & What is the budget for the project?\\
        & When does the project need to be completed?\\
        & What are the dimensions of the bridge?\\
        & What is the height of the bridge?\\
        & What is the clearance under the bridge?\\
        & What is the soil type in the area?\\
        & What is the water depth in the area?\\
        & What is the tide range in the area?\\
        & What is the wind speed in the area?\\
        & What is the temperature in the area?\\
        & What is the weather like in the area?\\
        & What is the traffic volume in the area?\\
        & What is the terrain in the area?\\
        & What is the geology in the area?\\
        & What is the topography in the area?\\
        & What is the vegetation in the area?\\
        & What is the land use in the area?\\
        & What are the restrictions in the area?\\
        & Is the area accessible?\\
        & What is the soil stability in the area?\\
        & What is the rock quality in the area?\\
        & What is the foundation depth in the area?\\
        & What is the water table in the area?\\
        & What is the water pressure in the area?\\
        & What is the seismic activity in the area?\\
        & What is the wind loading in the area?\\
        & What is the snow load in the area?\\
        & What is the ice load in the area?\\
        & What are the construction materials available in the area?\\
        & What is the construction method that will be used?\\
        & What is the estimated cost of the project?\\
        & When is the project scheduled to be completed?\\
        & What are the risks associated with the project?\\
        & What are the mitigation measures that will be taken?\\
        & What are the benefits of the project?\\
        & What are the drawbacks of the project?\\
        & What is the expected traffic volume?\\
        & What are the expected traffic patterns?\\
         \bottomrule \\
    \end{tabular}
    \caption{
        Generated questions for additional task-19 along with the prompt fed to GPT3.
    }
    \label{tab:additionalquestions19}
\end{table*}

\begin{table*}
    \centering
    \begin{tabular}{p{1.7cm}p{12.5 cm}}
        \toprule
        element & taskname/input prompt/model predictions  \\
        \midrule
        task & hospital setup planning\\
        \midrule
        prompt & I am a famous planner with a lot of experience in setting up hospitals. I will ask some questions to collect information and then I will make a perfect plan setting up a hospital in your area.\\
        \midrule
        questions& What is the current state of healthcare in your area?\\
        & What are the biggest challenges that need to be addressed in terms of healthcare in your area?\\
        & What is the population of your area?\\
        & What is the average income of residents in your area?\\
        & What is the land area of your area?\\
        & Is there a need for a new hospital in your area?\\
        & What type of hospital do you think would be most beneficial for your area?\\
        & What services do you think should be offered at the new hospital?\\
        & Do you have any ideas about where the new hospital should be located?\\
        & Do you have any ideas about how the new hospital should be funded?\\
        & Do you have any other ideas or suggestions about setting up a new hospital in your area?\\
         \bottomrule \\
    \end{tabular}
    \caption{
        Generated questions for additional task-20 along with the prompt fed to GPT3.
    }
    \label{tab:additionalquestions20}
\end{table*}

\begin{table*}
    \centering
    \begin{tabular}{p{1.7cm}p{12.5 cm}}
        \toprule
        element & taskname/input prompt/model predictions  \\
        \midrule
        task & water supply route setup\\
        \midrule
        prompt & I am a famous planner with a lot of experience in setting up water supply routes. I will ask some questions to collect information and then I will make a perfect plan for setting up water supply routes in your area.\\
        \midrule
        questions& What is the size of your area?\\
        & What is the population density of your area?\\
        & What is the average rainfall in your area?\\
        & What is the average temperature in your area?\\
        & What is the terrain like in your area?\\
        & What is the soil type in your area?\\
        & What is the available water resources in your area?\\
        & What is the demand for water in your area?\\
        & What are the restrictions on water use in your area?\\
        & What are the potential sources of pollution in your area?\\
        & What are the existing water treatment facilities in your area?\\
        & What is the capacity of the existing water treatment facilities in your area?\\
        & What are the existing sewer and stormwater infrastructure in your area?\\
        & What is the capacity of the existing sewer and stormwater infrastructure in your area?\\
        & What are the existing roads and transportation infrastructure in your area?\\
        & What is the existing power infrastructure in your area?\\
        & What are the existing communication infrastructure in your area?\\
        & What are the existing buildings and structures in your area?\\
        & What is the land use in your area?\\
        & What are the zoning regulations in your area?\\
        & What are the building codes in your area?\\
        & What are the existing environmental conditions in your area?\\
        & What are the existing social conditions in your area?\\
        & What is the existing economic conditions in your area?\\
        & What is the government support in your area?\\
        & What are the legal restrictions in your area?\\
        & What are the cultural restrictions in your area?\\
        & What are the financial restrictions in your area?\\
        & What are the infrastructure restrictions in your area?\\
        & What are the land use restrictions in your area?\\
        & What are the water use restrictions in your area?\\
        & What are the potential sources of new water supply in your area?\\
         \bottomrule \\
    \end{tabular}
    \caption{
        Generated questions for additional task-21 along with the prompt fed to GPT3.
    }
    \label{tab:additionalquestions21}
\end{table*}

\begin{table*}
    \centering
    \begin{tabular}{p{1.7cm}p{12.5 cm}}
        \toprule
        element & taskname/input prompt/model predictions  \\
        \midrule
        task & plan for running school\\
        \midrule
        prompt & I have a lot of experience in running schools. I am joining your school as the principal today. I will ask some questions to collect information and then I will make a perfect plan for running this school.\\
        \midrule
        questions& What are the main goals of the school?\\
        & What are the current methods for achieving these goals?\\
        & What are the challenges facing the school?\\
        & What resources does the school have available to achieve its goals?\\
        & What are the stakeholders in the school and what are their expectations?\\
        & What is the school's budget?\\
        & What are the school's fundraising goals?\\
        & What are the school's marketing goals?\\
        & What are the school's admissions goals?\\
        & What are the school's retention goals?\\
        & What are the school's graduation goals?\\
        & What are the school's post-graduation goals?\\
        & What are the school's accreditation goals?\\
        & What are the school's facilities goals?\\
        & What are the school's safety goals?\\
        & What are the school's technology goals?\\
        & What are the school's transportation goals?\\
        & What are the school's food service goals?\\
        & What are the school's extracurricular goals?\\
        & What are the school's community outreach goals?\\
        & What are the school's alumni relations goals?\\
        & What are the school's media relations goals?\\
        & What are the school's government relations goals?\\
        & What are the school's international relations goals?\\
        & What are the school's legal compliance goals?\\
        & What are the school's privacy goals?\\
        & What are the school's data security goals?\\
        & What are the school's emergency preparedness goals?\\
        & What are the school's sustainability goals?\\
        & What are the school's equity and inclusion goals?\\
        & What are the school's human resources goals?\\
        & What are the school's finance goals?\\
        & What are the school's information technology goals?\\
        & What are the school's facilities management goals?\\
        & What are the school's risk management goals?\\
        & What are the school's security goals?\\
        & What are the school's transportation goals?\\
        & What are the school's food service goals?\\
        & What are the school's environmental health and safety goals?\\
         \bottomrule \\
    \end{tabular}
    \caption{
        Generated questions for additional task-22 along with the prompt fed to GPT3.
    }
    \label{tab:additionalquestions22}
\end{table*}

\begin{table*}
    \centering
    \begin{tabular}{p{1.7cm}p{12.5 cm}}
        \toprule
        element & taskname/input prompt/model predictions  \\
        \midrule
        task & course curriculum design\\
        \midrule
        prompt & I am a famous course curriculum designer. I will ask some questions to collect information and then I will design a perfect curriculum for the course you want to teach.\\
        \midrule
        questions& What course do you want to teach?\\
        & What are your goals for the course?\\
        & What do you hope your students will gain from the course?\\
        & What are your requirements for the course?\\
        & What resources will you need for the course?\\
        & What is the format of the course?\\
        & What are the assessment methods for the course?\\
        & What are the grading criteria for the course?\\
        & What is the schedule for the course?\\
        & What are the policies for the course?\\
        & Do you have any other information that you think would be helpful?
\\
         \bottomrule \\
    \end{tabular}
    \caption{
        Generated questions for additional task-23 along with the prompt fed to GPT3.
    }
    \label{tab:additionalquestions23}
\end{table*}

\begin{table*}
    \centering
    \begin{tabular}{p{1.7cm}p{12.5 cm}}
        \toprule
        element & taskname/input prompt/model predictions  \\
        \midrule
        task & meeting agenda development\\
        \midrule
        prompt & I have a lot of experience in developing agenda for meetings. I will ask some questions to collect information and then I will design a perfect agenda for your meeting.\\
        \midrule
        questions& How long will the meeting be?\\
        & What is the purpose of the meeting?\\
        & Who will be attending the meeting?\\
        & What topics need to be covered in the meeting?\\
        & Are there any other specific instructions or requirements?\\
        & What is the format of the meeting?\\
        & What is the meeting location?\\
        & What is the date and time of the meeting?\\
        & Who will be chairing the meeting?\\
        & Who will be taking minutes?\\
        & Is there anything else you would like to add?\\
        & Would you like me to send you a draft agenda?\\
        & When do you need the agenda?\\
        & What is your email address?\\
        & What is your name?\\
        & What is the name of the project?\\
        & What is the project number?\\
        & What is the name of the client?\\
        & What is the name of the project manager?\\
        & What is the name of the company?\\
         \bottomrule \\
    \end{tabular}
    \caption{
        Generated questions for additional task-24 along with the prompt fed to GPT3.
    }
    \label{tab:additionalquestions24}
\end{table*}

\begin{table*}
    \centering
    \begin{tabular}{p{1.7cm}p{12.5 cm}}
        \toprule
        element & taskname/input prompt/model predictions  \\
        \midrule
        task & award acceptance speech preparation\\
        \midrule
        prompt & I have a lot of experience in writing award acceptance speech. I will ask some questions to collect information and then I will design a perfect award acceptance speech for you.\\
        \midrule
        questions& What is the occasion?\\
        & What is the award for?\\
        & What is the name of the award?\\
        & Who is the award being presented by?\\
        & What is your name?\\
        & What is the name of the movie you won the award for?\\
        & What are your thoughts on winning the award?\\
        & Who are you dedicating the award to?\\
        & What is your speech going to be about?\\
        & What is the tone of your speech?\\
        & What is the main point of your speech?\\
        & Are you going to thank anyone in your speech?\\
        & Is there anything else you would like to add?\\
        & Would you like me to write your speech for you?\\
        & When do you need the speech by?\\
        & Do you have any other instructions?\\
        & Would you like me to send you a draft of the speech before the final version?\\
        & What is your email address?\\
        & What is your phone number?\\
         \bottomrule \\
    \end{tabular}
    \caption{
        Generated questions for additional task-25 along with the prompt fed to GPT3.
    }
    \label{tab:additionalquestions25}
\end{table*}

\begin{table*}
    \centering
    \begin{tabular}{p{1.7cm}p{12.5 cm}}
        \toprule
        element & taskname/input prompt/model predictions  \\
        \midrule
        task & examination schedule preparation\\
        \midrule
        prompt & I am an expert in preparing examination schedule. I will ask some questions to collect information and then I will prepare an examination schedule for your school.\\
        \midrule
        questions& How many students are in your school?\\
        & How many classrooms do you have?\\
        & How many teachers do you have?\\
        & What is the average class size?\\
        & What is the length of each class period?\\
        & What is the number of class periods in a day?\\
        & What is the start time and end time of the school day?\\
        & How many days are in a school week?\\
        & What are the school holidays?\\
        & How many examinations do you have in a year?\\
        & What is the duration of each examination?\\
        & What is the maximum number of students that can be accommodated in a classroom?\\
         \bottomrule \\
    \end{tabular}
    \caption{
        Generated questions for additional task-26 along with the prompt fed to GPT3.
    }
    \label{tab:additionalquestions26}
\end{table*}

\begin{table*}
    \centering
    \begin{tabular}{p{1.7cm}p{12.5 cm}}
        \toprule
        element & taskname/input prompt/model predictions  \\
        \midrule
        task & news summarizing\\
        \midrule
        prompt & I am a famous news editor. I will ask some questions to collect information on the foreign policy of US and then I will summarize the articles for you.\\
        \midrule
        questions& What are the top three priorities of the United States when it comes to foreign policy?\\
        & What are the top three countries that the United States has diplomatic relations with?\\
        & What are the top three issues that the United States is currently facing in its foreign policy?\\
        & What are the top three ways that the United States has been involved in foreign policy in the past?\\
        & What are the top three ways that the United States can be involved in foreign policy in the future?\\
        & What are the top three benefits of the United States' current foreign policy?\\
        & What are the top three drawbacks of the United States' current foreign policy?\\
         \bottomrule \\
    \end{tabular}
    \caption{
        Generated questions for additional task-27 along with the prompt fed to GPT3.
    }
    \label{tab:additionalquestions27}
\end{table*}

\begin{table*}
    \centering
    \begin{tabular}{p{1.7cm}p{12.5 cm}}
        \toprule
        element & taskname/input prompt/model predictions  \\
        \midrule
        task & news channel recommendation\\
        \midrule
        prompt & I have an experience in watching news on various channels. I will ask some questions to collect information and then I will suggest which news source is appropriate for you.\\
        \midrule
        questions& How frequently do you watch news?\\
        & How much time do you spend watching news every day?\\
        & What kind of news do you prefer?\\
        & What is your favorite news source?\\
        & What do you think of the news sources available to you?\\
        & Do you have any concerns about the news you are watching?\\
        & What do you think of the objectivity of the news you are watching?\\
        & What do you think of the fairness of the news you are watching?\\
        & Do you think the news you are watching is representative of the diversity of opinions in the world?\\
        & Do you think the news you are watching is reflective of your own values and beliefs?\\
         \bottomrule \\
    \end{tabular}
    \caption{
        Generated questions for additional task-28 along with the prompt fed to GPT3.
    }
    \label{tab:additionalquestions28}
\end{table*}

\begin{table*}
    \centering
    \begin{tabular}{p{1.7cm}p{12.5 cm}}
        \toprule
        element & taskname/input prompt/model predictions  \\
        \midrule
        task & news article generation\\
        \midrule
        prompt & I am a famous journalist. I will ask some questions to collect information on the event happened with you and then I will write a nice article for you.\\
        \midrule
        questions& What is your name?\\
        & What is your age?\\
        & What is the name of the event?\\
        & When did the event happen?\\
        & Where did the event happen?\\
        & What was the purpose of the event?\\
        & Who organized the event?\\
        & How many people participated in the event?\\
        & What was your experience at the event?\\
        & Would you recommend the event to others?\\
         \bottomrule \\
    \end{tabular}
    \caption{
        Generated questions for additional task-29 along with the prompt fed to GPT3.
    }
    \label{tab:additionalquestions29}
\end{table*}

\begin{table*}
    \centering
    \begin{tabular}{p{1.7cm}p{12.5 cm}}
        \toprule
        element & taskname/input prompt/model predictions  \\
        \midrule
        task & post match presentation interview (with the wining team captain)\\
        \midrule
        prompt & I am a famous sports journalist. Congratulations on winning today's match. I am going to interview you as part of the post-match presentation ceremony.\\
        \midrule
        questions& What do you think was the key to your success today?\\
        & What was your game plan coming into today's match?\\
        & What do you think the team can improve on going forward?\\
        & What was your favorite part of today's match?\\
        & What are your thoughts on your opponent's performance today?\\
        & What's next for you and the team?\\
         \bottomrule \\
    \end{tabular}
    \caption{
        Generated questions for additional task-30 along with the prompt fed to GPT3.
    }
    \label{tab:additionalquestions30}
\end{table*}

\begin{table*}
    \centering
    \begin{tabular}{p{1.7cm}p{12.5 cm}}
        \toprule
        element & taskname/input prompt/model predictions  \\
        \midrule
        task & post match press-conference (with the losing team captain)\\
        \midrule
        prompt & I am a famous sports journalist. Tough luck for you today. I am going to interview you as part of the post-match press conference.\\
        \midrule
        questions& How do you think you played today?\\
        & What do you think was the key to the game?\\
        & What are your thoughts on the team's performance?\\
        & What do you think the team needs to work on?\\
        & What was your favorite part of the game?\\
        & What are your thoughts on the fans?\\
        & Do you have any final thoughts?\\
         \bottomrule \\
    \end{tabular}
    \caption{
        Generated questions for additional task-31 along with the prompt fed to GPT3.
    }
    \label{tab:additionalquestions31}
\end{table*}

\begin{table*}
    \centering
    \begin{tabular}{p{1.7cm}p{12.5 cm}}
        \toprule
        element & taskname/input prompt/model predictions  \\
        \midrule
        task & customized history snippet generation\\
        \midrule
        prompt & I am a famous historian. I will ask some questions to collect information on the situation you are in and then I will generate a history snippet for you.\\
        \midrule
        questions& Who are you?\\
        & What is the situation you are in?\\
        & What year is it?\\
        & What is your favorite color?\\
        & In your opinion, what is the most important event in history?\\
        & Why do you think that is the most important event in history?\\
        & What is your favorite historical moment?\\
        & Why is that your favorite historical moment?\\
        &  If you could go back in time and witness any event, what would it be?\\
        & What do you think the future holds for humanity?\\
        & What is your favorite thing about history?\\
        & What do you think is the most important lesson that history can teach us?\\
        & If you could change one thing about history, what would it be?\\
        & What do you think is the most significant event of the 21st century?\\
        & What do you think will be the most important event of the 22nd century?\\
        & What is your favorite historical figure?\\
         \bottomrule \\
    \end{tabular}
    \caption{
        Generated questions for additional task-32 along with the prompt fed to GPT3.
    }
    \label{tab:additionalquestions32}
\end{table*}

\begin{table*}
    \centering
    \begin{tabular}{p{1.7cm}p{12.5 cm}}
        \toprule
        element & taskname/input prompt/model predictions  \\
        \midrule
        task & political speech generation\\
        \midrule
        prompt & I am an expert in writing political speeches. I will ask some questions to collect information and then I will write an awesome speech for you.\\
        \midrule
        questions& What is the main purpose of the speech?\\
        & What are the main points that you would like to communicate?\\
        & Is there anything else you would like me to know?\\
        & Who is the audience for this speech?\\
        & What is the occasion for this speech?\\
        & Where will the speech be given?\\
        & When will the speech be given?\\
        & What is the tone of the speech?\\
        & What is the length of the speech?\\
        & Are there any specific things you would like me to mention?\\
         \bottomrule \\
    \end{tabular}
    \caption{
        Generated questions for additional task-33 along with the prompt fed to GPT3.
    }
    \label{tab:additionalquestions33}
\end{table*}

\begin{table*}
    \centering
    \begin{tabular}{p{1.7cm}p{12.5 cm}}
        \toprule
        element & taskname/input prompt/model predictions  \\
        \midrule
        task & political manifesto generation\\
        \midrule
        prompt & I am an expert in creating manifesto for politicians. I will ask some questions to collect information and then I will create a manifesto for you.\\
        \midrule
        questions& What are your top three priorities if elected?\\
        & What are your thoughts on taxation?\\
        & What are your thoughts on education?\\
        & What are your thoughts on healthcare?\\
        & What are your thoughts on the environment?\\
        & What are your thoughts on crime?\\
        & What are your thoughts on the economy?\\
        & What are your thoughts on infrastructure?\\
        & What are your thoughts on government spending?\\
        & What are your thoughts on national security?\\
        & What are your thoughts on immigration?\\
        & What are your thoughts on gun control?\\
        & What are your thoughts on foreign policy?\\
        & What are your thoughts on the budget deficit?\\
        & What are your thoughts on jobs and the economy?\\
         \bottomrule \\
    \end{tabular}
    \caption{
        Generated questions for additional task-34 along with the prompt fed to GPT3.
    }
    \label{tab:additionalquestions34}
\end{table*}

\begin{table*}
    \centering
    \begin{tabular}{p{1.7cm}p{12.5 cm}}
        \toprule
        element & taskname/input prompt/model predictions  \\
        \midrule
        task & managing election campaign\\
        \midrule
        prompt & I am an expert in managing election campaigns. I will ask some questions to collect information and then I will design an election campaign for you.\\
        \midrule
        questions& What is your budget for this campaign?\\
        & Who is your target audience?\\
        & What are your campaign goals?\\
        & What is your timeline for this campaign?\\
        & What type of election is this?\\
        & What is the geographical scope of your campaign?\\
        & What type of media do you want to use for your campaign?\\
        & What is your message?\\
        & What are your key issues?\\
        & Who is your competition?\\
        & What are your strengths?\\
        & What are your weaknesses?\\
        & What are your risks?\\
        & What are your opportunities?\\
        & What are your threats?\\
        & What is your plan B?\\
        & What is your contingency plan?\\
        & Do you have any additional information that you would like to share?\\
         \bottomrule \\
    \end{tabular}
    \caption{
        Generated questions for additional task-35 along with the prompt fed to GPT3.
    }
    \label{tab:additionalquestions35}
\end{table*}

\begin{table*}
    \centering
    \begin{tabular}{p{1.7cm}p{12.5 cm}}
        \toprule
        element & taskname/input prompt/model predictions  \\
        \midrule
        task & lawyer for a land dispute case\\
        \midrule
        prompt & I am a famous lawyer. Sorry to hear about your land dispute case. I will ask some questions to collect information necessary to fight your case.\\
        \midrule
        questions& What are the facts of your case?\\
        & Who is your opponent in the case?\\
        & What is your legal argument?\\
        & What is the relief that you are seeking?\\
        & What is your opponent's legal argument?\\
        & What is the relief that your opponent is seeking?\\
        & What is the status of the case?\\
        & What is the most recent development in the case?\\
        & What is the next scheduled hearing in the case?\\
        & What is your trial strategy?\\
        & What is your opponent's trial strategy?\\
        & What is the likely outcome of the case?\\
        & What are the risks of going to trial?\\
        & What are the benefits of going to trial?\\
        & What are the risks of settling the case?\\
        & What are the benefits of settling the case?\\
        & What is your bottom line settlement demand?\\
        & What is your opponent's bottom line settlement demand?\\
        & What are the chances of settlement?\\
        & What is the most likely outcome if the case goes to trial?\\
        & What are the risks and benefits of each option?\\
        & What is your recommendation?\\
         \bottomrule \\
    \end{tabular}
    \caption{
        Generated questions for additional task-36 along with the prompt fed to GPT3.
    }
    \label{tab:additionalquestions36}
\end{table*}

\begin{table*}
    \centering
    \begin{tabular}{p{1.7cm}p{12.5 cm}}
        \toprule
        element & taskname/input prompt/model predictions  \\
        \midrule
        task & flight refund application\\
        \midrule
        prompt & I have a lot of experience in getting refunds for flight delays. I will ask some questions to collect information and then I will contact the airline authority on your behalf requesting refund.\\
        \midrule
        questions& What was the airline?\\
        & What was the flight number?\\
        & What was the date of the flight?\\
        & What was the origin of the flight?\\
        & What was the destination of the flight?\\
        & What was the scheduled departure time of the flight?\\
        & What was the actual departure time of the flight?\\
        & What was the scheduled arrival time of the flight?\\
        & What was the actual arrival time of the flight?\\
        & How long was the flight delayed?\\
         \bottomrule \\
    \end{tabular}
    \caption{
        Generated questions for additional task-37 along with the prompt fed to GPT3.
    }
    \label{tab:additionalquestions37}
\end{table*}

\begin{table*}
    \centering
    \begin{tabular}{p{1.7cm}p{12.5 cm}}
        \toprule
        element & taskname/input prompt/model predictions  \\
        \midrule
        task & movie story customization\\
        \midrule
        prompt & I am a famous storyteller. I specialize in editing popular movie stories and customizing them for users. I will ask some questions and edit storyline of the movie 'Titanic' based on your answers.\\
        \midrule
        questions& How would you like the story to end?\\
        & What kind of ending would you like?\\
        & What other changes would you like to see in the story?\\
        & Would you like to see the two main characters end up together?\\
        & Is there anything else you would like to change about the story?\\
         \bottomrule \\
    \end{tabular}
    \caption{
        Generated questions for additional task-38 along with the prompt fed to GPT3.
    }
    \label{tab:additionalquestions38}
\end{table*}

\begin{table*}
    \centering
    \begin{tabular}{p{1.7cm}p{12.5 cm}}
        \toprule
        element & taskname/input prompt/model predictions  \\
        \midrule
        task & music composition\\
        \midrule
        prompt & I am a famous music composer. I will ask some questions to collect your music requirements and then I will compose a song specially for you.\\
        \midrule
        questions& What is the occasion for which you need the song?\\
        & What is the mood or feeling that you would like the song to convey?\\
        & What specific elements would you like the song to include?\\
        & Would you like the song to be in a particular style or genre?\\
        & Is there anything else you would like me to know about your song request?\\
         \bottomrule \\
    \end{tabular}
    \caption{
        Generated questions for additional task-39 along with the prompt fed to GPT3.
    }
    \label{tab:additionalquestions39}
\end{table*}

\begin{table*}
    \centering
    \begin{tabular}{p{1.7cm}p{12.5 cm}}
        \toprule
        element & taskname/input prompt/model predictions  \\
        \midrule
        task & movie transcript generation\\
        \midrule
        prompt & I am an expert in generating movie transcripts. I will ask some questions to collect your music requirements and then I will write transcript for you.\\
        \midrule
        questions& What kind of movie are you looking for a transcript for?\\
        & What is the name of the movie?\\
        & What is the plot of the movie?\\
        & What is the genre of the movie?\\
        & What is the budget for the movie?\\
        & What is the release date for the movie?\\
        & What is the running time for the movie?\\
        & Who are the cast members for the movie?\\
        & Who is the director of the movie?\\
        & Who is the producer of the movie?\\
        & Who is the writer of the movie?\\
         \bottomrule \\
    \end{tabular}
    \caption{
        Generated questions for additional task-40 along with the prompt fed to GPT3.
    }
    \label{tab:additionalquestions40}
\end{table*}

\begin{table*}
    \centering
    \begin{tabular}{p{1.7cm}p{12.5 cm}}
        \toprule
        element & taskname/input prompt/model predictions  \\
        \midrule
        task & novel publication\\
        \midrule
        prompt & I am a novel publisher. I will ask some questions to collect information about your novel and then tell if I can publish it.\\
        \midrule
        questions& What is the title of your novel?\\
        & What is the genre of your novel?\\
        & What is the plot of your novel?\\
        & What is the target audience for your novel?\\
        & Is your novel complete?\\
        & What is the word count of your novel?\\
        & Do you have a marketing plan for your novel?\\
        & Do you have any publishing experience?\\
        & Do you have any literary agent representation?\\
        & Do you have a platform?\\
        & Do you have any writing awards?\\
        & Do you have any previous publications?\\
        & What are your writing credentials?\\
        & What are your professional credentials?\\
        & Do you have a personal website?\\
        & Do you have a blog?\\
        & Do you have an active social media presence?\\
        & Do you have any other professional experience?\\
        & Why do you think your novel would be a good fit for my publishing company?\\
        & Do you have any questions for me?\\
         \bottomrule \\
    \end{tabular}
    \caption{
        Generated questions for additional task-41 along with the prompt fed to GPT3.
    }
    \label{tab:additionalquestions41}
\end{table*}

\begin{table*}
    \centering
    \begin{tabular}{p{1.7cm}p{12.5 cm}}
        \toprule
        element & taskname/input prompt/model predictions  \\
        \midrule
        task & food preparation\\
        \midrule
        prompt & I am a famous chef. I will ask some questions to collect your requirements and then I will prepare a dish of your choice.\\
        \midrule
        questions& What is your name?\\
        & What is your favorite type of food?\\
        & What are your dietary restrictions?\\
        & What is your favorite type of cuisine?\\
        & What are your favorite dishes?\\
        & What ingredients do you not like?\\
        & Is there anything else you would like me to know?\\
         \bottomrule \\
    \end{tabular}
    \caption{
        Generated questions for additional task-42 along with the prompt fed to GPT3.
    }
    \label{tab:additionalquestions42}
\end{table*}

\begin{table*}
    \centering
    \begin{tabular}{p{1.7cm}p{12.5 cm}}
        \toprule
        element & taskname/input prompt/model predictions  \\
        \midrule
        task & restaurant menu preparation\\
        \midrule
        prompt & I am an expert in running restaurants. I will ask some questions to collect your requirements and then I will prepare a restaurant menu of your choice.\\
        \midrule
        questions& What type of restaurant menu do you want?\\
        & What type of food do you want to serve?\\
        & What is your budget?\\
        & What are your dietary restrictions?\\
        & What is your target audience?\\
        & What is your restaurant's theme?\\
        & What is the occasion?\\
        & What is the atmosphere of your restaurant?\\
        & What is the location of your restaurant?\\
        & What is the ambiance of your restaurant?\\
        & What is the service style of your restaurant?\\
        & What is the dress code of your restaurant?\\
        & What is the price range of your restaurant?\\
        & What is the cuisine of your restaurant?\\
        & What is the name of your restaurant?\\
        & What is the seating capacity of your restaurant?\\
        & What is the contact information of your restaurant?\\
        & What are the hours of operation of your restaurant?\\
        & What are the days of operation of your restaurant?\\
        & What is the speciality of your restaurant?\\
        & What is the history of your restaurant?\\ 
        & What are the awards of your restaurant?\\
        & What is the website of your restaurant?\\
        & What is the Facebook page of your restaurant?\\
        & What is the Instagram of your restaurant?\\
        & What is the Twitter of your restaurant?\\
         \bottomrule \\
    \end{tabular}
    \caption{
        Generated questions for additional task-43 along with the prompt fed to GPT3.
    }
    \label{tab:additionalquestions43}
\end{table*}

\begin{table*}
    \centering
    \begin{tabular}{p{1.7cm}p{12.5 cm}}
        \toprule
        element & taskname/input prompt/model predictions  \\
        \midrule
        task & diet plan\\
        \midrule
        prompt & I am a famous dietitian. I will ask some questions to collect information and then I will suggest diet plan for you.\\
        \midrule
        questions& What is your weight?\\
        & What is your height?\\
        & What is your activity level?\\
        & What are your food preferences?\\
        & What are your health concerns?\\
        & What medications are you currently taking?\\
        & Are you trying to lose, gain, or maintain your weight?\\
        & How many meals per day do you typically eat?\\
        & How many snacks per day do you typically eat?\\
         \bottomrule \\
    \end{tabular}
    \caption{
        Generated questions for additional task-44 along with the prompt fed to GPT3.
    }
    \label{tab:additionalquestions44}
\end{table*}

\begin{table*}
    \centering
    \footnotesize 
    \begin{tabular}{p{1.7cm}p{12.5 cm}}
        \toprule
        element & taskname/input prompt/model predictions  \\
        \midrule
        task & physics problem solving\\
        \midrule
        prompt & I am a famous physicist. I am here to help you solve the physics problem you have on mechanics. I will ask some questions in this process.\\
        \midrule
        questions& What is the physics problem you have on mechanics?\\
        & What are the objects involved in the problem?\\
        & What are the forces acting on the objects?\\
        & What is the goal of the problem?\\
        & What are the constraints of the problem?\\
        & What is the starting configuration of the problem?\\
        & What is the ending configuration of the problem?\\
        & What is the time frame of the problem?\\
        & What are the units of measurement for the problem?\\
        & Is there any friction in the problem?\\
        & Are the objects in the problem moving in a straight line or in a circle?\\
        & What is the mass of the objects in the problem?\\
        & What is the velocity of the objects in the problem?\\
        & What is the acceleration of the objects in the problem?\\
        & What is the resulting force on the objects in the problem?\\
        & What is the resulting torque on the objects in the problem?\\
        & What is the resulting energy on the objects in the problem?\\
        & What is the resulting momentum on the objects in the problem?\\
        & What are the conditions for equilibrium in the problem?\\
        & What is the period of the problem?\\
        & What is the frequency of the problem?\\
        & What is the wavelength of the problem?\\
        & What is the wave function of the problem?\\
        & What is the amplitude of the problem?\\
        & What is the phase of the problem?\\
        & What is the wave vector of the problem?\\
        & What is the group velocity of the problem?\\
        & What is the phase velocity of the problem?\\
        & What is the energy of the problem?\\
        & What is the momentum of the problem?\\
        & What is the position of the objects in the problem?\\
        & What is the displacement of the objects in the problem?\\
        & What is the distance traveled by the objects in the problem?\\
        & What is the time of the problem?\\
        & What is the speed of the objects in the problem?\\
        & What is the power of the problem?\\
        & What is the kinetic energy of the objects in the problem?\\
        & What is the potential energy of the objects in the problem?\\
        & What is the total energy of the objects in the problem?\\
        & What is the angular momentum of the objects in the problem?\\
        & What is the angular velocity of the objects in the problem?\\
        & What is the angular acceleration of the objects in the problem?\\
        & What is the moment of inertia of the objects in the problem?\\
        & What is the rotational kinetic energy of the objects in the problem?\\
        & What is the gravitational potential energy of the objects in the problem?\\
        & What is the elastic potential energy of the objects in the problem?\\
        & What is the magnetic potential energy of the objects in the problem?\\
        & What is the electric potential energy of the objects in the problem?\\
        & What is the nuclear potential energy of the objects in the problem?\\
        & What is the thermal energy of the objects in the problem?\\
        & What is the chemical energy of the objects in the problem?\\
         \bottomrule \\
    \end{tabular}
    \caption{
        Generated questions for additional task-45 along with the prompt fed to GPT3.
    }
    \label{tab:additionalquestions45}
\end{table*}

\begin{table*}
    \centering
    \begin{tabular}{p{1.7cm}p{12.5 cm}}
        \toprule
        element & taskname/input prompt/model predictions  \\
        \midrule
        task & chemistry problem solving\\
        \midrule
        prompt & I am a famous chemistry teacher. I am here to help you find the appropriate chemical equation to solve the chemistry problem you have on stoichiometry. I will now ask clarifying questions to better understand your problem.\\
        \midrule
        questions& What is the problem you are trying to solve?\\
        & What is the molarity of the acid?\\
        & What is the volume of the acid?\\
        & What is the molarity of the base?\\
        & What is the volume of the base?\\
        & What is the desired molarity of the solution?\\
        & What is the desired volume of the solution?\\
         \bottomrule \\
    \end{tabular}
    \caption{
        Generated questions for additional task-46 along with the prompt fed to GPT3.
    }
    \label{tab:additionalquestions46}
\end{table*}

\begin{table*}
    \centering
    \begin{tabular}{p{1.7cm}p{12.5 cm}}
        \toprule
        element & taskname/input prompt/model predictions  \\
        \midrule
        task & painting house\\
        \midrule
        prompt & I have a lot of experience in painting houses. I will ask questions to collect information and then I will make a plan for painting your house.\\
        \midrule
        questions& What is the size of your house?\\
        & What is the condition of your house?\\
        & What is your budget for painting your house?\\
        & What is your timeline for painting your house?\\
        & What is your preferred color scheme for painting your house?\\
        & What is your preferred type of paint for painting your house?\\
        & Are you planning on painting your house yourself or hiring a professional?\\
        & Do you have any other specific requests or instructions for painting your house?\\
         \bottomrule \\
    \end{tabular}
    \caption{
        Generated questions for additional task-47 along with the prompt fed to GPT3.
    }
    \label{tab:additionalquestions47}
\end{table*}

\begin{table*}
    \centering
    \begin{tabular}{p{1.7cm}p{12.5 cm}}
        \toprule
        element & taskname/input prompt/model predictions  \\
        \midrule
        task & car mechanic\\
        \midrule
        prompt & I am a famous car mechanic. I will ask questions to collect information and then I will tell you what needs to be fixed in your car.\\
        \midrule
        questions& What seems to be the problem with your car?\\
        & When did you first notice the problem?\\
        & Have you noticed the problem happening more frequently?\\
        & Is the problem happening all the time or only intermittently?\\
        & What are the circumstances under which the problem occurs?\\
        & What have you done to try to fix the problem?\\
        & What other information can you tell me about the problem?\\
         \bottomrule \\
    \end{tabular}
    \caption{
        Generated questions for additional task-48 along with the prompt fed to GPT3.
    }
    \label{tab:additionalquestions48}
\end{table*}


\clearpage

\begin{table*}
    \centering
    \begin{tabular}{p{1.7cm}p{12.5 cm}}
        \toprule
        element & taskname/input prompt/model predictions  \\
        \midrule
        task & assistance in filling bank deposit form\\
        \midrule
        prompt &I have a lot of experience in banking and have helped people in filling forms. I will ask questions to collect information and then I will fill the bank deposit form for you.\\
        \midrule
        questions& What is your name?\\
        & What is the name of your bank?\\
        & What is the account number?\\
        & What is the amount of money you would like to deposit?\\
        & Do you have a check or cash?\\
        & What is the date?\\
        & Is this a savings or checking account?\\
        & Would you like a receipt?\\
         \bottomrule \\
    \end{tabular}
    \caption{
        Generated questions for additional task-49 along with the prompt fed to GPT3.
    }
    \label{tab:additionalquestions49}
\end{table*}

\begin{table*}
    \centering
    \begin{tabular}{p{1.7cm}p{12.5 cm}}
        \toprule
        element & taskname/input prompt/model predictions  \\
        \midrule
        task & electricity bill payment\\
        \midrule
        prompt &I have a lot of experience in paying electricity bills online. I will ask questions to collect information and then I will pay the electricity bill for you.\\
        \midrule
        questions& What is the account number for the electricity bill?\\
        & What is the amount of the electricity bill?\\
        & What is the due date for the electricity bill?\\
        & What is the website for the electricity company?\\
        & What is your username for the electricity company website?\\
        & What is your password for the electricity company website?\\
        & What is your email address?\\
        & What is your phone number?\\
        & What is your full name?\\
        & What is the name of the electricity company?\\
         \bottomrule \\
    \end{tabular}
    \caption{
        Generated questions for additional task-50 along with the prompt fed to GPT3.
    }
    \label{tab:additionalquestions50}
\end{table*}

\begin{table*}
    \centering
    \begin{tabular}{p{1.7cm}p{12.5 cm}}
        \toprule
        element & taskname/input prompt/model predictions  \\
        \midrule
        task & creating todo list\\
        \midrule
        prompt &I have a lot of experience in generating ToDo list. I will ask questions to collect information and then I will create a todo list for you.\\
        \midrule
        questions& What is the goal of the todo list?\\
        & What are the main tasks that need to be completed in order to achieve the goal?\\
        & What are the deadlines for each task?\\
        & Who is responsible for each task?\\
        & What is the priority level for each task?\\
        & Is there anything else that needs to be taken into account when creating the todo list?\\
        & Are there any dependencies between tasks?\\
        & What is the expected outcome of the todo list?\\
        & What are the risks and potential roadblocks associated with the todo list?\\
        & What are the possible ways to overcome the risks and roadblocks?\\
        & Are there any other considerations that need to be taken into account?\\
         \bottomrule \\
    \end{tabular}
    \caption{
        Generated questions for additional task-51 along with the prompt fed to GPT3.
    }
    \label{tab:additionalquestions51}
\end{table*}

\begin{table*}
    \centering
    \begin{tabular}{p{1.7cm}p{12.5 cm}}
        \toprule
        element & taskname/input prompt/model predictions  \\
        \midrule
        task & creating post travel report\\
        \midrule
        prompt &I have a lot of experience in generating post travel report. I will ask questions to collect information and then I will prepare your post travel report.\\
        \midrule
        questions& How did you find the overall experience?\\
        & What were the good points of the trip?\\
        & What were the bad points of the trip?\\
        & How much did you spend in total?\\
        & Would you recommend this trip to others? Why or why not?\\
        & Is there anything you would have done differently?\\
        & What was your favorite part of the trip?\\
        & What was your least favorite part of the trip?\\
        & How was the accommodation?\\
        & How was the food?\\
        & How were the locals?\\
        & Did you feel safe during the trip?\\
        & Would you like to visit this destination again? Why or why not?\\
        & Is there anything else you would like to add?\\
         \bottomrule \\
    \end{tabular}
    \caption{
        Generated questions for additional task-52 along with the prompt fed to GPT3.
    }
    \label{tab:additionalquestions52}
\end{table*}

\begin{table*}
    \centering
    \begin{tabular}{p{1.7cm}p{12.5 cm}}
        \toprule
        element & taskname/input prompt/model predictions  \\
        \midrule
        task & creating child monitoring report\\
        \midrule
        prompt &I have a lot of experience in monitoring child. I will ask questions to collect information and then I will prepare your child-monitoring report.\\
        \midrule
        questions& How often do you check in on your child when they are home alone?\\
        & How do you feel when you are away from your child and they are home alone?\\
        & What worries you the most about your child being home alone?\\
        & What have you done to ensure your child knows what to do if there is an emergency while you are away?\\
        & What are your child’s favorite activities to do when they are home alone?\\
        & Has your child ever had any problems while home alone? If so, what happened and how did they handle it?\\
        & Do you have any other concerns or questions about your child being home alone?\\
         \bottomrule \\
    \end{tabular}
    \caption{
        Generated questions for additional task-53 along with the prompt fed to GPT3.
    }
    \label{tab:additionalquestions53}
\end{table*}

\begin{table*}
    \centering
    \begin{tabular}{p{1.7cm}p{12.5 cm}}
        \toprule
        element & taskname/input prompt/model predictions  \\
        \midrule
        task & treating mental illness\\
        \midrule
        prompt &I have a lot of experience in handling mental problem. I will ask questions to collect information and then I will suggest a treatment for you.\\
        \midrule
        questions& What is your name?\\
        & What is your age?\\
        & What is your occupation?\\
        & What are your hobbies?\\
        & Do you have any medical conditions?\\
        & Do you have any mental health conditions?\\
        & Do you have any history of mental illness in your family?\\
        & What is your current mental state?\\
        & What are your current stressors?\\
        & What are your current coping mechanisms?\\
        & What are your current support systems?\\
        & What are your current treatment options?\\
        & What are your goals for treatment?\\
        & What are your thoughts on medication?\\
        & What are your thoughts on therapy?\\
        & What are your thoughts on self-care?\\
        & What are your thoughts on support groups?\\
        & What are your thoughts on lifestyle changes?\\
        & What are your thoughts on relapse prevention?\\
        & What are your thoughts on long-term treatment?\\
        & What are your thoughts on medication compliance?\\
        & What are your thoughts on therapy attendance?\\
        & What are your thoughts on self-care consistency?\\
        & What are your thoughts on support group participation?\\
         \bottomrule \\
    \end{tabular}
    \caption{
        Generated questions for additional task-54 along with the prompt fed to GPT3.
    }
    \label{tab:additionalquestions54}
\end{table*}

\begin{table*}
    \centering
    \begin{tabular}{p{1.7cm}p{12.5 cm}}
        \toprule
        element & taskname/input prompt/model predictions  \\
        \midrule
        task & career counselling\\
        \midrule
        prompt &I am an expert in career counselling. I will ask questions to collect information and then I will prepare a suitable career plan for you.\\
        \midrule
        questions& What are your long-term career goals?\\
        & What are your educational qualifications?\\
        & What is your work experience?\\
        & What are your skills and strengths?\\
        & What are your interests?\\
        & What are your values?\\
        & What are your goals for the next five years?\\
        & What are your plans for further education or training?\\
        & What are your plans for your personal life?\\
        & What are your plans for your financial future?\\
        & What are your plans for your retirement?\\
        & What other factors are important to you in choosing a career?\\
        & What are your thoughts on changing careers?\\
        & What are your thoughts on starting your own business?\\
        & What are your thoughts on working for someone else?\\
        & What are your thoughts on working abroad?\\
        & What are your thoughts on taking a break from your career?\\
        & What other questions do you have about your career?\\
        & Are you ready to start working on your career plan?\\
        & Do you have any concerns or questions that you would like to discuss before we start?\\
        & Are you sure that you are ready to commit to this career plan?\\
        & Would you like me to review your career plan with you on a regular basis?\\
        & Is there anything else that you would like to discuss about your career plan?\\
         \bottomrule \\
    \end{tabular}
    \caption{
        Generated questions for additional task-55 along with the prompt fed to GPT3.
    }
    \label{tab:additionalquestions55}
\end{table*}

\begin{table*}
    \centering
    \begin{tabular}{p{1.7cm}p{12.5 cm}}
        \toprule
        element & taskname/input prompt/model predictions  \\
        \midrule
        task & child tutoring\\
        \midrule
        prompt &I am an expert in child tutoring I will ask questions to collect information and then I will decide if I can tutor your child.\\
        \midrule
        questions& What is your child's name?\\
        & What is your child's grade level?\\
        & What are the subjects your child needs help with?\\
        & How often would you like tutoring sessions to be?\\
        & What days and times are best for you?\\
        & What is your child's learning style?\\
        & Does your child have any special needs?\\
        & Is your child behind, at grade level, or ahead in school?\\
        & How do you feel about your child's progress in school?\\
        & What are your goals for your child's education?\\
        & What are your child's strengths and weaknesses?\\
        & Is there anything else you feel I should know about your child?\\
        & Would you like me to assess your child before we begin tutoring?\\
        & Do you have any questions for me?\\
        & When would you like to start?\\
         \bottomrule \\
    \end{tabular}
    \caption{
        Generated questions for additional task-56 along with the prompt fed to GPT3.
    }
    \label{tab:additionalquestions56}
\end{table*}

\begin{table*}
    \centering
    \begin{tabular}{p{1.7cm}p{12.5 cm}}
        \toprule
        element & taskname/input prompt/model predictions  \\
        \midrule
        task & paper writing\\
        \midrule
        prompt &I am a famous researcher with a lot of experience in writing papers. I will ask questions to collect information about your research and then I will suggest a plan to write your paper.\\
        \midrule
        questions& What is the topic of your research?\\
        & How much experience do you have with writing papers?\\
        & What is your writing style?\\
        & What is the format of your paper?\\
        & What is the deadline for your paper?\\
        & What is the word count for your paper?\\
        & What is the purpose of your paper?\\
        & Who is your audience?\\
        & What are your expectations for this project?\\
        & What is your budget for this project?\\
        & What is the timeline for this project?\\
        & What are the deliverables for this project?\\
        & What are your research methods?\\
        & What sources will you be using for your research?\\
        & What are your thoughts on the research process?\\
        & What are your thoughts on writing papers?\\
         \bottomrule \\
    \end{tabular}
    \caption{
        Generated questions for additional task-57 along with the prompt fed to GPT3.
    }
    \label{tab:additionalquestions57}
\end{table*}


\end{document}